\PassOptionsToPackage{table,dvipsnames}{xcolor}
\documentclass[11pt,a4paper]{berkeley}
\usepackage{probeagentstyle}
\probeagentpreprinttrue
\usepackage[numbers,sort&compress]{natbib}
\usepackage{fontspec}
\usepackage[scale=0.93]{plex-mono}
\usepackage{unicode-math}
\usepackage{probeagentpreprint}

\newcommand{\FigFunnel}{%
\begin{tikzpicture}[x=1cm,y=1cm,
    lgd/.style={anchor=west,font=\sffamily\footnotesize,text=soft}]
  \useasboundingbox (-0.05,-5.32) rectangle (13.72,0.30);
  \def\P{0.335}                       
  \def\W{30}                          
  \def\LX{10.55}                      

  \node[anchor=west,font=\sffamily\bfseries\small,text=ink] at (0.10,-0.30) {a)};
  \node[lgd] at (0.52,-0.30) {239 failure explanations, one square each};
  \foreach \i in {0,...,238}{
    \pgfmathtruncatemacro{\rw}{\i/\W}
    \pgfmathtruncatemacro{\cl}{mod(\i,\W)}
    \pgfmathsetmacro{\xx}{0.10+\cl*\P}
    \pgfmathsetmacro{\yy}{-0.95-\rw*\P}
    \ifnum\i<119 \wcell{\xx}{\yy}{pri!85}\else \wcell{\xx}{\yy}{hair!70}\fi}
  \node[lgd] at (\LX,-0.80) {\sw{pri!85}\figval{119} put to siblings};
  \node[lgd] at (\LX,-1.24) {\sw{hair!70}\figval{120} never queued};
  \node[lgd,anchor=north west,align=left] at (\LX,-1.58)
    {\figsub{129 exploration nodes}\\[-1pt]\figsub{318 verification nodes}\\[-1pt]%
     \figsub{bounded search budget}\\[-1pt]\figsub{for sibling construction}};

  \node[anchor=west,font=\sffamily\bfseries\small,text=ink] at (0.10,-3.62) {b)};
  \node[lgd] at (0.52,-3.62) {their verdicts, one square each};
  \foreach \i in {0,...,118}{
    \pgfmathtruncatemacro{\rw}{\i/\W}
    \pgfmathtruncatemacro{\cl}{mod(\i,\W)}
    \pgfmathsetmacro{\xx}{0.10+\cl*\P}
    \pgfmathsetmacro{\yy}{-4.15-\rw*\P}
    \ifnum\i<20 \wcell{\xx}{\yy}{pri}
    \else\ifnum\i<46 \wcell{\xx}{\yy}{cat5}
    \else\ifnum\i<94 \wcell{\xx}{\yy}{cat3}
    \else \wcell{\xx}{\yy}{hair!70}\fi\fi\fi}
  \node[lgd] at (\LX,-4.00) {\sw{pri}\figval{20} agent behavior};
  \node[lgd] at (\LX,-4.38) {\sw{cat5}\figval{26} weak};
  \node[lgd] at (\LX,-4.76) {\sw{cat3}\figval{48} task-specific};
  \node[lgd] at (\LX,-5.14) {\sw{hair!70}\figval{25} unverified};
\end{tikzpicture}}

\newcommand{\FigGranularity}{%
\begin{tikzpicture}[x=1cm,y=1cm,
    ttl/.style={anchor=west,font=\sffamily\footnotesize,text=soft},
    num/.style={font=\sffamily\scriptsize,text=ink},
    tik/.style={font=\sffamily\scriptsize,text=soft}]
  \useasboundingbox (-0.05,-7.05) rectangle (13.70,0.30);
  \def\X{1.55}\def\S{1.62}            
  \def\Ba{-3.10}                      
  \pgfmathsetmacro{\H}{2.30/700}      

  \node[anchor=west,font=\sffamily\bfseries\small,text=ink] at (0.10,-0.28) {a)};
  \node[ttl] at (0.52,-0.28) {induced groups and multi-failure support};
  \node[ttl] at (1.28,-0.78) {\sw{pri!25}all groups\quad\sw{pri!85}$\geq$2 failures};

  \foreach \i/\lab/\tot/\multi in {0/off/23/17, 1/{$0$}/374/168, 2/{$.02$}/421/155,
      3/{$.05$}/505/124, 4/{$.08$}/565/101, 5/{$.12$}/603/76, 6/{$.20$}/665/38}{
    \pgfmathsetmacro{\xx}{\X+\i*\S}
    \fill[pri!25,rounded corners=0.6pt]
      ({\xx-0.34},\Ba) rectangle ({\xx+0.34},{\Ba+\tot*\H});
    \fill[pri!85,rounded corners=0.6pt]
      ({\xx-0.34},\Ba) rectangle ({\xx+0.34},{\Ba+\multi*\H});
    \node[num,anchor=south] at (\xx,{\Ba+0.05+\tot*\H}) {\tot};
    \node[tik,anchor=north] at (\xx,{\Ba-0.09}) {\lab};}
  \draw[dwire] ({\X-0.60},\Ba) -- (12.35,\Ba);
  \node[tik,anchor=north] at ({\X+3*\S},-3.56) {density margin $\mu$};

  \node[font=\sffamily\scriptsize,text=pri,align=center,anchor=south] at (\X,-2.72)
    {one group holds\\[-2pt]650 of 961 failures};
  \draw[dwire] (\X,-2.74) -- (\X,-3.06);

  \node[anchor=west,font=\sffamily\bfseries\small,text=ink] at (0.10,-4.15) {b)};
  \node[ttl] at (0.52,-4.15) {held-out membership agreement};

  \def\B{-6.55}
  \pgfmathsetmacro{\A}{1.35/0.30}     
  \fill[pri!7] ({\X+0.5*\S},{\B-0.07}) rectangle ({\X+4.5*\S},-4.90);
  \node[font=\sffamily\scriptsize\bfseries,text=hdr,anchor=south] at ({\X+2.5*\S},{\B+0.09})
    {intermediate margins};
  \foreach \g in {0.70,0.80,0.90}{
    \pgfmathsetmacro{\gy}{\B+(\g-0.66)*\A}
    \draw[hair!60,line width=0.4pt] ({\X-0.60},\gy) -- (12.35,\gy);
    \node[tik,anchor=east] at ({\X-0.66},\gy) {\g};}
  \draw[dwire] ({\X-0.60},\B) -- (12.35,\B);

  \draw[cat6,line width=1.0pt]
    \foreach \i/\acc [count=\n] in {0/0.707, 1/0.874, 2/0.902, 3/0.914, 4/0.921, 5/0.926,
        6/0.931}{\ifnum\n>1 -- \fi ({\X+\i*\S},{\B+(\acc-0.66)*\A})};
  \foreach \i/\acc in {0/0.707, 1/0.874, 2/0.902, 3/0.914, 4/0.921, 5/0.926, 6/0.931}{
    \pgfmathsetmacro{\xx}{\X+\i*\S}
    \pgfmathsetmacro{\yy}{\B+(\acc-0.66)*\A}
    \fill[white] (\xx,\yy) circle (0.075);
    \fill[cat6] (\xx,\yy) circle (0.055);
    \node[tik,anchor=south] at (\xx,{\yy+0.11}) {\acc};}

  \node[font=\sffamily\scriptsize,text=pri,align=right,anchor=east] at (12.35,-6.16)
    {94\% of groups are singletons;\\[-2pt]none spans two agents};
\end{tikzpicture}}

\newcommand{\FigReplay}{%
\begin{tikzpicture}[x=1cm,y=1cm]
  \useasboundingbox (-0.05,-7.62) rectangle (13.7,0.30);
  \def\L{4.55}          
  \def\Wd{7.00}         
  \pgfmathsetmacro{\U}{\Wd/1.0}

  \foreach \t in {0,0.25,0.5,0.75,1.0}{
    \pgfmathsetmacro{\xx}{\L+\t*\U}
    \draw[hair,line width=0.4pt] (\xx,-0.42) -- (\xx,-6.62);
    \pgfmathsetmacro{\pct}{\t*100}
    \node[font=\rmfamily\scriptsize,text=soft,anchor=north] at (\xx,-6.66)
      {\pgfmathprintnumber[fixed,precision=0]{\pct}\%};}

  \node[font=\rmfamily\scriptsize,text=soft,anchor=south] at ({\L+0.5*\U},-0.40)
    {matched failure rate and 95\% Wilson interval};

  \foreach \i/\est/\lo/\hi/\name/\col/\lg/\vd in {%
     0/0.222/0.090/0.452/{Terminus 2 / Kimi K3}/pri/tb/ms,
     1/0.278/0.125/0.509/{Codex / GPT-5.6-sol}/cat6/cx/oa,
     2/0.333/0.163/0.563/{Terminus 2 / GLM-5}/pri/tb/za,
     3/0.389/0.203/0.614/{Terminus 2 / GPT-5.6-sol}/pri/tb/oa,
     4/0.444/0.246/0.663/{Terminus 2 / Claude Opus 5}/pri/tb/cl,
     5/0.500/0.290/0.710/{Terminus 2 / GLM-5.2}/pri/tb/za,
     6/0.500/0.290/0.710/{Terminus 2 / Claude Sonnet 4.5}/pri/tb/cl,
     7/0.556/0.337/0.754/{Terminus 2 / Qwen3.8-Max}/pri/tb/qw,
     8/0.667/0.437/0.837/{Terminus 2 / DeepSeek-V4}/pri/tb/ds,
     9/0.667/0.437/0.837/{Terminus 2 / MiniMax-M2.7}/pri/tb/mm,
    10/0.667/0.437/0.837/{Terminus 2 / Claude Sonnet 5}/pri/tb/cl,
    11/0.722/0.491/0.875/{Terminus 2 / Grok 4.5}/pri/tb/xa,
    12/0.722/0.491/0.875/{Terminus 2 / Claude Opus 4.8}/pri/tb/cl,
    13/0.778/0.548/0.910/{Claude Code / Claude Sonnet 4.5}/cat6/cc/cl,
    14/0.944/0.742/0.990/{Terminus 2 / GPT-5.4-nano}/pri/tb/oa}{
      \pgfmathsetmacro{\yy}{-0.62-\i*0.375}
      \node[anchor=east,font=\rmfamily\scriptsize,text=body] at ({\L-0.18},\yy)
        {\agmark{\lg}{\vd}\name};
      \interval{\yy}{\L+\est*\U}{\L+\lo*\U}{\L+\hi*\U}{\col}
      \pgfmathsetmacro{\epct}{\est*100}
      \node[anchor=west,font=\rmfamily\scriptsize,text=soft] at ({\L+\hi*\U+0.12},\yy)
        {\pgfmathprintnumber[fixed,precision=1]{\epct}\%};}

  \draw[cat3,line width=0.7pt]
    ({\L+\Wd+1.05},-2.12) -- ({\L+\Wd+1.20},-2.12)
    -- ({\L+\Wd+1.20},-5.12) -- ({\L+\Wd+1.05},-5.12);
  \node[rotate=-90,font=\rmfamily\scriptsize,text=cat3,align=center]
    at ({\L+\Wd+1.62},-3.62) {Opus pair\\[-2pt]{\footnotesize\color{soft}6:1, $p=0.13$}};

  \node[anchor=north,font=\rmfamily\footnotesize,text=soft] at ({\L+0.5*\Wd},-6.98)
    {failure rate on the shared discriminating pool, 18 rollouts each, 95\% Wilson};
  \node[anchor=north,font=\rmfamily\footnotesize,text=soft] at ({\L+0.5*\Wd},-7.30)
    {\sw{pri}Terminus 2 \quad \sw{cat6}other harness};
\end{tikzpicture}}

\newcommand{\FigTaxonomy}{%
\begin{tikzpicture}[x=1cm,y=1cm]
  \useasboundingbox (-0.05,-4.30) rectangle (13.70,0.42);
  \def\L{3.10}\def\BW{3.20}
  \pgfmathsetmacro{\U}{\BW/163}
  \pgfmathsetmacro{\Ma}{\L+\BW+0.95}   
  \pgfmathsetmacro{\Mb}{\L+\BW+1.95}   
  \node[anchor=east,font=\sffamily\scriptsize,text=soft] at (\Ma,0.16) {groups};
  \node[anchor=east,font=\sffamily\scriptsize,text=soft] at (\Mb,0.16) {failures};
  \foreach \i/\name/\modes/\recs in {%
    0/{\ttfamily incomplete}/163/264, 1/{\ttfamily instruction-violation}/107/179,
    2/{\ttfamily verification-failure}/102/165, 3/{\ttfamily tool-misuse}/51/100,
    4/{\ttfamily reasoning-error}/47/88, 5/{\ttfamily planning}/45/68,
    6/{\ttfamily hallucination}/44/67, 7/{\ttfamily environment-error}/17/25}{
      \pgfmathsetmacro{\yy}{-0.55-\i*0.42}
      \node[anchor=east,font=\scriptsize,text=body] at ({\L-0.16},\yy) {\name};
      \fill[pri!16,rounded corners=0.6pt] (\L,{\yy-0.135}) rectangle ({\L+\BW},{\yy+0.135});
      \fill[pri!82,rounded corners=0.6pt] (\L,{\yy-0.135}) rectangle ({\L+\modes*\U},{\yy+0.135});
      \node[anchor=east,font=\sffamily\scriptsize,text=ink] at (\Ma,\yy) {\modes};
      \node[anchor=east,font=\sffamily\scriptsize,text=soft] at (\Mb,\yy) {\recs};}
  \node[anchor=west,font=\sffamily\footnotesize,text=soft] at (\L,-4.02)
    {\sw{pri!82}groups in this category\quad\sw{pri!16}scaled to the largest, 163};

  \def\TA{8.45}\def\TW{5.20}
  \pgfmathsetmacro{\Cb}{\TA+3.70}      
  \pgfmathsetmacro{\Cc}{\TA+\TW-0.14}  
  \fill[band] (\TA,-0.76) rectangle ({\TA+\TW},-0.36);
  \node[anchor=east,font=\sffamily\scriptsize\bfseries,text=hdr] at ({\Cb-0.48},-0.56) {fixed taxonomy};
  \node[anchor=east,font=\sffamily\scriptsize\bfseries,text=hdr] at (\Cc,-0.56) {\textsc{ProbeAgent}};
  \foreach \i/\lab/\ta/\tb in {%
      0/{buckets}/{8}/{505},
      1/{largest, failures}/{264}/{11},
      2/{largest, share}/{27.5\%}/{1.1\%},
      3/{median, failures}/{94}/{1},
      4/{buckets of one failure}/{0}/{381},
      5/{failures with no bucket}/{5}/{0}}{
    \pgfmathsetmacro{\yy}{-0.98-\i*0.42}
    \ifodd\i\else\fill[zebra] (\TA,{\yy-0.21}) rectangle ({\TA+\TW},{\yy+0.21});\fi
    \node[anchor=west,font=\sffamily\scriptsize,text=body] at ({\TA+0.14},\yy) {\lab};
    \node[anchor=east,font=\sffamily\scriptsize,text=ink] at (\Cb,\yy) {\ta};
    \node[anchor=east,font=\sffamily\scriptsize,text=ink] at (\Cc,\yy) {\tb};}
\end{tikzpicture}}

\newcommand{\FigCalibration}{%
\begin{tikzpicture}[x=1cm,y=1cm]
  \useasboundingbox (-0.05,-1.82) rectangle (13.7,0.85);
  \def\L{0.70}\def\Wd{12.4}
  \pgfmathsetmacro{\U}{\Wd/2.0}      

  \fill[band] (\L,-0.28) rectangle ({\L+\Wd},0.06);
  \foreach \v in {-1,-0.5,0,0.5,1}{
    \pgfmathsetmacro{\xx}{\L+(\v+1)*\U}
    \draw[hair,line width=0.4pt] (\xx,-0.34) -- (\xx,-0.28);
    \node[font=\sffamily\scriptsize,text=soft,anchor=north] at (\xx,-0.36)
      {\pgfmathprintnumber[fixed,precision=1]{\v}};}

  \pgfmathsetmacro{\fa}{\L+(-0.10+1)*\U}
  \pgfmathsetmacro{\fb}{\L+(0.15+1)*\U}
  \fill[cat3!45] (\fa,-0.28) rectangle (\fb,0.06);
  \draw[cat3,line width=0.7pt] (\fa,-0.28) rectangle (\fb,0.06);
  \node[anchor=north,font=\sffamily\scriptsize,text=cat3,align=center] at ({(\fa+\fb)/2},-0.74)
    {fixed default $[-0.10,\ 0.15]$ \quad \textbf{9\%} judged ``same''};

  \pgfmathsetmacro{\ca}{\L+(-0.03+1)*\U}
  \pgfmathsetmacro{\cb}{\L+(0.64+1)*\U}
  \fill[cat2!30] (\ca,0.14) rectangle (\cb,0.48);
  \draw[cat2,line width=0.7pt] (\ca,0.14) rectangle (\cb,0.48);
  \node[anchor=south,font=\sffamily\scriptsize,text=cat2,align=center] at ({(\ca+\cb)/2},0.52)
    {judge-anchored $[-0.03,\ 0.64]$ \quad \textbf{41\%} judged ``same''};

  \node[anchor=north,font=\sffamily\footnotesize,text=soft] at (7.0,-1.32)
    {centred cosine between two failure failures};
  \node[anchor=north west,font=\sffamily\scriptsize,text=soft,align=left] at (0.10,-1.02)
    {settled as\\[-2pt]different};
  \node[anchor=north east,font=\sffamily\scriptsize,text=soft,align=right] at (13.60,-1.02)
    {settled as\\[-2pt]same};
\end{tikzpicture}}

\newcommand{\FigCost}{%
\begin{tikzpicture}[x=1cm,y=1cm,
    tik/.style={font=\sffamily\scriptsize,text=soft},
    pin/.style={font=\sffamily\scriptsize,text=body}]
  \useasboundingbox (-0.05,-8.92) rectangle (13.70,0.30);
  \def\L{1.70}\def\Wd{11.90}
  \def\B{-7.80}\def\Ht{7.15}
  \def\xa{5.0}\def\xb{6.6}          

  \foreach \d/\lab in {5/{100k}, 5.30103/{200k}, 5.69897/{500k}, 6/{1M}, 6.30103/{2M}}{
    \pgfmathsetmacro{\xx}{\L+(\d-\xa)/(\xb-\xa)*\Wd}
    \draw[hair!60,line width=0.4pt] (\xx,\B) -- (\xx,{\B+\Ht});
    \node[tik,anchor=north] at (\xx,{\B-0.08}) {\lab};}
  \foreach \r in {0,25,50,75,100}{
    \pgfmathsetmacro{\yy}{\B+\r/100*\Ht}
    \draw[hair!60,line width=0.4pt] (\L,\yy) -- ({\L+\Wd},\yy);
    \node[tik,anchor=east] at ({\L-0.10},\yy) {\r\%};}
  \draw[dwire] (\L,\B) -- ({\L+\Wd},\B);
  \node[tik,anchor=north] at ({\L+0.5*\Wd},-8.22)
    {input tokens per rollout (log scale)};
  \node[tik,rotate=90] at (0.24,{\B+0.5*\Ht}) {failure rate on the shared pool};

  \foreach \lx/\rate/\vd/\ring in {%
      5.0479/66.7/cl/pri, 5.0732/22.2/ms/pri, 5.1173/38.9/oa/pri, 5.1665/55.6/qw/pri,
      5.3890/50.0/za/pri, 5.5167/72.2/xa/pri, 5.5308/33.3/za/pri, 5.5626/44.4/cl/pri,
      5.6215/66.7/mm/pri, 5.6307/66.7/ds/pri, 5.6730/50.0/cl/pri, 5.9235/94.4/oa/pri,
      6.0519/27.8/oa/cat6, 6.0620/72.2/cl/pri, 6.5427/77.8/cl/cat6}{
    \pgfmathsetmacro{\xx}{\L+(\lx-\xa)/(\xb-\xa)*\Wd}
    \pgfmathsetmacro{\yy}{\B+\rate/100*\Ht}
    \fill[white] (\xx,\yy) circle (0.185);
    \draw[\ring,line width=0.6pt] (\xx,\yy) circle (0.185);
    \node[inner sep=0pt] at (\xx,\yy) {\vlogoat{0.235cm}{\vd}};}

  \foreach \lx/\rate/\side/\nm in {%
      5.0479/66.7/r/{Claude Sonnet 5},
      5.0732/22.2/r/{Kimi K3},
      5.1173/38.9/r/{GPT-5.6-sol},
      5.1665/55.6/r/{Qwen3.8-Max},
      5.3890/50.0/r/{GLM-5.2},
      5.5167/72.2/l/{Grok 4.5},
      5.5308/33.3/r/{GLM-5},
      5.5626/44.4/r/{Claude Opus 5},
      5.6730/50.0/r/{Claude Sonnet 4.5},
      5.9235/94.4/r/{GPT-5.4-nano},
      6.0519/27.8/r/{GPT-5.6-sol \emph{under Codex}},
      6.0620/72.2/r/{Claude Opus 4.8},
      6.5427/77.8/l/{Claude Sonnet 4.5 \emph{under Claude Code}}}{
    \pgfmathsetmacro{\xx}{\L+(\lx-\xa)/(\xb-\xa)*\Wd}
    \pgfmathsetmacro{\yy}{\B+\rate/100*\Ht}
    \if\side r
      \node[pin,anchor=west] at ({\xx+0.26},\yy) {\nm};
    \else
      \node[pin,anchor=east] at ({\xx-0.26},\yy) {\nm};
    \fi}
  \node[pin,anchor=west] at ({\L+(5.6307-\xa)/(\xb-\xa)*\Wd+0.26},{\B+0.667*\Ht})
    {MiniMax-M2.7, DeepSeek-V4};

  \node[anchor=north,font=\sffamily\scriptsize,text=soft] at ({\L+0.5*\Wd},-8.56)
    {\tikz[baseline=-0.05ex]{\draw[pri,line width=0.6pt] (0,0) circle (0.09);}\;Terminus 2
     \qquad
     \tikz[baseline=-0.05ex]{\draw[cat6,line width=0.6pt] (0,0) circle (0.09);}\;Claude Code or Codex};

  \node[anchor=north west,font=\sffamily\footnotesize,text=soft,align=left,
        text width=11.60cm,inner sep=0pt] at ({\L+0.10},-6.58)
    {\figval{$\rho = 0.50$}\;\figsub{Spearman correlation across 13 Terminus 2 configurations
     ($p = 0.08$).}};
\end{tikzpicture}}

\newcommand{\FigPairs}{%
\begin{tikzpicture}[x=1cm,y=1cm,
    tik/.style={font=\sffamily\scriptsize,text=soft},
    val/.style={font=\sffamily\scriptsize,text=ink}]
  \useasboundingbox (-0.05,-3.52) rectangle (13.70,0.30);
  \def\Wd{3.30}                      
  \pgfmathsetmacro{\U}{\Wd/70}
  \def\PA{3.05}\def\PB{10.30}        

  \foreach \P/\ttl/\lt in {\PA/{same family, newer version}/a, \PB/{same model, other harness}/b}{
    \foreach \v in {0,35,70}{
      \pgfmathsetmacro{\xx}{\P+\v*\U}
      \draw[hair!60,line width=0.4pt] (\xx,-2.66) -- (\xx,-0.52);
      \node[tik,anchor=north] at (\xx,-2.70) {\v};}
    \node[tik,anchor=north] at ({\P+0.5*\Wd},-3.04) {distinct failure groups};}
  \node[anchor=west,font=\sffamily\bfseries\small,text=ink] at (0.10,-0.15) {a)};
  \node[anchor=west,font=\sffamily\footnotesize,text=soft] at (0.52,-0.15)
    {same model family, newer version};
  \node[anchor=west,font=\sffamily\bfseries\small,text=ink] at (7.35,-0.15) {b)};
  \node[anchor=west,font=\sffamily\footnotesize,text=soft] at (7.77,-0.15)
    {same model, different harness};

  \foreach \i/\vd/\na/\va/\nb/\vb in {%
      0/cl/{Sonnet 4.5}/52/{Sonnet 5}/56,
      1/za/{GLM-5}/35/{GLM-5.2}/6,
      2/cl/{Opus 4.8}/8/{Opus 5}/24}{
    \pgfmathsetmacro{\yy}{-1.00-\i*0.62}
    \pgfmathsetmacro{\xa}{\PA+\va*\U}\pgfmathsetmacro{\xb}{\PA+\vb*\U}
    \node[anchor=east,font=\sffamily\scriptsize,text=body] at ({\PA-0.20},\yy)
      {\agmark{tb}{\vd}\na\,$\rightarrow$\,\nb};
    \draw[hair,line width=2.2pt] (\xa,\yy) -- (\xb,\yy);
    \fill[cat5] (\xa,\yy) circle (0.085);
    \node[val,anchor=south] at (\xa,{\yy+0.11}) {\va};
    \fill[pri] (\xb,\yy) circle (0.095);
    \node[val,anchor=north] at (\xb,{\yy-0.11}) {\vb};}

  \foreach \i/\vd/\sa/\va/\vb/\nm in {%
      0/cl/cc/16/52/{Claude Sonnet 4.5},
      1/oa/cx/7/58/{GPT-5.6-sol}}{
    \pgfmathsetmacro{\yy}{-1.00-\i*0.62}
    \pgfmathsetmacro{\xa}{\PB+\va*\U}\pgfmathsetmacro{\xb}{\PB+\vb*\U}
    \node[anchor=east,font=\sffamily\scriptsize,text=body] at ({\PB-0.20},\yy)
      {\agarrow{\sa}{tb}\nm};
    \draw[hair,line width=2.2pt] (\xa,\yy) -- (\xb,\yy);
    \fill[cat5] (\xa,\yy) circle (0.085);
    \node[val,anchor=south] at (\xa,{\yy+0.11}) {\va};
    \fill[pri] (\xb,\yy) circle (0.095);
    \node[val,anchor=south] at (\xb,{\yy+0.11}) {\vb};}
\end{tikzpicture}}

\newcommand{\FigPairShare}{%
\begin{tikzpicture}[x=1cm,y=1cm]
  \useasboundingbox (-0.05,-4.13) rectangle (13.70,0.42);
  \node[anchor=east,font=\sffamily\small,text=ink,align=right,text width=3.0cm] at (3.3299999999999996,-0.14) {the 9 pairs\\we designed};
  \wcell{3.55}{-0.28}{pri}
  \wcell{3.89}{-0.28}{pri}
  \wcell{4.23}{-0.28}{pri}
  \wcell{4.57}{-0.28}{pri}
  \wcell{4.91}{-0.28}{hair!70}
  \wcell{5.25}{-0.28}{hair!70}
  \wcell{5.59}{-0.28}{hair!70}
  \wcell{5.93}{-0.28}{hair!70}
  \wcell{6.27}{-0.28}{hair!70}
  \node[anchor=west,font=\sffamily\small,text=pri] at (6.829999999999999,-0.14) {4 of 9 share a failure group};
  \node[anchor=east,font=\sffamily\small,text=ink,align=right,text width=3.0cm] at (3.3299999999999996,-1.46) {every other\\pair in the panel};
  \wcell{3.55}{-1.24}{pri}
  \wcell{3.89}{-1.24}{pri}
  \wcell{4.23}{-1.24}{pri}
  \wcell{4.57}{-1.24}{pri}
  \wcell{4.91}{-1.24}{pri}
  \wcell{5.25}{-1.24}{pri}
  \wcell{5.59}{-1.24}{pri}
  \wcell{5.93}{-1.24}{pri}
  \wcell{6.27}{-1.24}{pri}
  \wcell{6.61}{-1.24}{pri}
  \wcell{6.95}{-1.24}{pri}
  \wcell{7.29}{-1.24}{pri}
  \wcell{7.63}{-1.24}{pri}
  \wcell{7.97}{-1.24}{pri}
  \wcell{8.31}{-1.24}{pri}
  \wcell{8.65}{-1.24}{pri}
  \wcell{8.99}{-1.24}{pri}
  \wcell{9.33}{-1.24}{pri}
  \wcell{9.67}{-1.24}{pri}
  \wcell{10.01}{-1.24}{pri}
  \wcell{10.35}{-1.24}{pri}
  \wcell{10.69}{-1.24}{pri}
  \wcell{11.03}{-1.24}{pri}
  \wcell{11.37}{-1.24}{pri}
  \wcell{11.71}{-1.24}{pri}
  \wcell{12.05}{-1.24}{pri}
  \wcell{3.55}{-1.58}{hair!70}
  \wcell{3.89}{-1.58}{hair!70}
  \wcell{4.23}{-1.58}{hair!70}
  \wcell{4.57}{-1.58}{hair!70}
  \wcell{4.91}{-1.58}{hair!70}
  \wcell{5.25}{-1.58}{hair!70}
  \wcell{5.59}{-1.58}{hair!70}
  \wcell{5.93}{-1.58}{hair!70}
  \wcell{6.27}{-1.58}{hair!70}
  \wcell{6.61}{-1.58}{hair!70}
  \wcell{6.95}{-1.58}{hair!70}
  \wcell{7.29}{-1.58}{hair!70}
  \wcell{7.63}{-1.58}{hair!70}
  \wcell{7.97}{-1.58}{hair!70}
  \wcell{8.31}{-1.58}{hair!70}
  \wcell{8.65}{-1.58}{hair!70}
  \wcell{8.99}{-1.58}{hair!70}
  \wcell{9.33}{-1.58}{hair!70}
  \wcell{9.67}{-1.58}{hair!70}
  \wcell{10.01}{-1.58}{hair!70}
  \wcell{10.35}{-1.58}{hair!70}
  \wcell{10.69}{-1.58}{hair!70}
  \wcell{11.03}{-1.58}{hair!70}
  \wcell{11.37}{-1.58}{hair!70}
  \wcell{11.71}{-1.58}{hair!70}
  \wcell{12.05}{-1.58}{hair!70}
  \wcell{3.55}{-1.92}{hair!70}
  \wcell{3.89}{-1.92}{hair!70}
  \wcell{4.23}{-1.92}{hair!70}
  \wcell{4.57}{-1.92}{hair!70}
  \wcell{4.91}{-1.92}{hair!70}
  \wcell{5.25}{-1.92}{hair!70}
  \wcell{5.59}{-1.92}{hair!70}
  \wcell{5.93}{-1.92}{hair!70}
  \wcell{6.27}{-1.92}{hair!70}
  \wcell{6.61}{-1.92}{hair!70}
  \wcell{6.95}{-1.92}{hair!70}
  \wcell{7.29}{-1.92}{hair!70}
  \wcell{7.63}{-1.92}{hair!70}
  \wcell{7.97}{-1.92}{hair!70}
  \wcell{8.31}{-1.92}{hair!70}
  \wcell{8.65}{-1.92}{hair!70}
  \wcell{8.99}{-1.92}{hair!70}
  \wcell{9.33}{-1.92}{hair!70}
  \wcell{9.67}{-1.92}{hair!70}
  \wcell{10.01}{-1.92}{hair!70}
  \wcell{10.35}{-1.92}{hair!70}
  \wcell{10.69}{-1.92}{hair!70}
  \wcell{11.03}{-1.92}{hair!70}
  \wcell{11.37}{-1.92}{hair!70}
  \wcell{11.71}{-1.92}{hair!70}
  \wcell{12.05}{-1.92}{hair!70}
  \wcell{3.55}{-2.26}{hair!70}
  \wcell{3.89}{-2.26}{hair!70}
  \wcell{4.23}{-2.26}{hair!70}
  \wcell{4.57}{-2.26}{hair!70}
  \wcell{4.91}{-2.26}{hair!70}
  \wcell{5.25}{-2.26}{hair!70}
  \wcell{5.59}{-2.26}{hair!70}
  \wcell{5.93}{-2.26}{hair!70}
  \wcell{6.27}{-2.26}{hair!70}
  \wcell{6.61}{-2.26}{hair!70}
  \wcell{6.95}{-2.26}{hair!70}
  \wcell{7.29}{-2.26}{hair!70}
  \wcell{7.63}{-2.26}{hair!70}
  \wcell{7.97}{-2.26}{hair!70}
  \wcell{8.31}{-2.26}{hair!70}
  \wcell{8.65}{-2.26}{hair!70}
  \wcell{8.99}{-2.26}{hair!70}
  \wcell{9.33}{-2.26}{hair!70}
  \wcell{9.67}{-2.26}{hair!70}
  \wcell{10.01}{-2.26}{hair!70}
  \wcell{10.35}{-2.26}{hair!70}
  \wcell{10.69}{-2.26}{hair!70}
  \wcell{11.03}{-2.26}{hair!70}
  \wcell{11.37}{-2.26}{hair!70}
  \wcell{11.71}{-2.26}{hair!70}
  \wcell{12.05}{-2.26}{hair!70}
  \wcell{3.55}{-2.60}{hair!70}
  \wcell{3.89}{-2.60}{hair!70}
  \wcell{4.23}{-2.60}{hair!70}
  \wcell{4.57}{-2.60}{hair!70}
  \wcell{4.91}{-2.60}{hair!70}
  \wcell{5.25}{-2.60}{hair!70}
  \wcell{5.59}{-2.60}{hair!70}
  \wcell{5.93}{-2.60}{hair!70}
  \wcell{6.27}{-2.60}{hair!70}
  \wcell{6.61}{-2.60}{hair!70}
  \wcell{6.95}{-2.60}{hair!70}
  \wcell{7.29}{-2.60}{hair!70}
  \wcell{7.63}{-2.60}{hair!70}
  \wcell{7.97}{-2.60}{hair!70}
  \wcell{8.31}{-2.60}{hair!70}
  \wcell{8.65}{-2.60}{hair!70}
  \wcell{8.99}{-2.60}{hair!70}
  \wcell{9.33}{-2.60}{hair!70}
  \wcell{9.67}{-2.60}{hair!70}
  \wcell{10.01}{-2.60}{hair!70}
  \wcell{10.35}{-2.60}{hair!70}
  \wcell{10.69}{-2.60}{hair!70}
  \wcell{11.03}{-2.60}{hair!70}
  \wcell{11.37}{-2.60}{hair!70}
  \wcell{11.71}{-2.60}{hair!70}
  \wcell{12.05}{-2.60}{hair!70}
  \wcell{3.55}{-2.94}{hair!70}
  \wcell{3.89}{-2.94}{hair!70}
  \wcell{4.23}{-2.94}{hair!70}
  \wcell{4.57}{-2.94}{hair!70}
  \wcell{4.91}{-2.94}{hair!70}
  \wcell{5.25}{-2.94}{hair!70}
  \wcell{5.59}{-2.94}{hair!70}
  \wcell{5.93}{-2.94}{hair!70}
  \wcell{6.27}{-2.94}{hair!70}
  \wcell{6.61}{-2.94}{hair!70}
  \wcell{6.95}{-2.94}{hair!70}
  \wcell{7.29}{-2.94}{hair!70}
  \wcell{7.63}{-2.94}{hair!70}
  \wcell{7.97}{-2.94}{hair!70}
  \wcell{8.31}{-2.94}{hair!70}
  \wcell{8.65}{-2.94}{hair!70}
  \wcell{8.99}{-2.94}{hair!70}
  \wcell{9.33}{-2.94}{hair!70}
  \wcell{9.67}{-2.94}{hair!70}
  \wcell{10.01}{-2.94}{hair!70}
  \wcell{10.35}{-2.94}{hair!70}
  \wcell{10.69}{-2.94}{hair!70}
  \wcell{11.03}{-2.94}{hair!70}
  \wcell{11.37}{-2.94}{hair!70}
  \wcell{11.71}{-2.94}{hair!70}
  \wcell{12.05}{-2.94}{hair!70}
  \wcell{3.55}{-3.28}{hair!70}
  \wcell{3.89}{-3.28}{hair!70}
  \wcell{4.23}{-3.28}{hair!70}
  \wcell{4.57}{-3.28}{hair!70}
  \wcell{4.91}{-3.28}{hair!70}
  \wcell{5.25}{-3.28}{hair!70}
  \wcell{5.59}{-3.28}{hair!70}
  \wcell{5.93}{-3.28}{hair!70}
  \wcell{6.27}{-3.28}{hair!70}
  \wcell{6.61}{-3.28}{hair!70}
  \wcell{6.95}{-3.28}{hair!70}
  \wcell{7.29}{-3.28}{hair!70}
  \wcell{7.63}{-3.28}{hair!70}
  \wcell{7.97}{-3.28}{hair!70}
  \wcell{8.31}{-3.28}{hair!70}
  \wcell{8.65}{-3.28}{hair!70}
  \wcell{8.99}{-3.28}{hair!70}
  \wcell{9.33}{-3.28}{hair!70}
  \wcell{9.67}{-3.28}{hair!70}
  \wcell{10.01}{-3.28}{hair!70}
  \wcell{10.35}{-3.28}{hair!70}
  \wcell{10.69}{-3.28}{hair!70}
  \wcell{11.03}{-3.28}{hair!70}
  \wcell{11.37}{-3.28}{hair!70}
  \wcell{11.71}{-3.28}{hair!70}
  \node[anchor=west,font=\sffamily\small,text=soft] at (12.610000000000001,-1.46) {26 of 181};
  \node[anchor=west,font=\sffamily\footnotesize,text=soft] at (0.0,-3.78) {\sw{pri}the two share at least one failure mode\quad\sw{hair!70}they share none};
\end{tikzpicture}}

\tikzset{
  ckTitle/.style={font=\sffamily\bfseries\fontsize{8.4}{10}\selectfont,text=ckInk,inner sep=0pt},
  ckSub/.style={font=\sffamily\fontsize{7}{8.5}\selectfont,text=ckMuted,inner sep=0pt},
  ckHead/.style={font=\sffamily\bfseries\fontsize{7}{8.5}\selectfont,text=ckText,inner sep=0pt},
  ckLabel/.style={font=\sffamily\fontsize{7.4}{9}\selectfont,text=ckText,inner sep=0pt},
  ckNum/.style={font=\sffamily\fontsize{7.4}{9}\selectfont,text=ckText,inner sep=0pt},
  ckValue/.style={font=\sffamily\bfseries\fontsize{7.4}{9}\selectfont,text=ckInk,inner sep=0pt},
  ckTick/.style={font=\sffamily\fontsize{6.6}{8}\selectfont,text=ckMuted,inner sep=0pt},
  ckNote/.style={font=\sffamily\fontsize{6.6}{8}\selectfont,text=ckMuted,inner sep=0pt},
  ckGrid/.style={draw=ckTrack,line width=0.45pt},
  ckBase/.style={draw=ckRule,line width=0.6pt},
  ckHeadRule/.style={draw=ckInk,line width=0.55pt},
  ckEndRule/.style={draw=ckRule,line width=0.4pt},
  ckLink/.style={draw=ckRule,line width=0.45pt,dash pattern=on 1pt off 1pt},
  ckStep/.style={draw=ckGrey,line width=0.45pt}
}
\newcommand{\cksq}[1]{\tikz[baseline=-0.05ex]\fill[#1] (0,0) rectangle (0.62em,0.62em);\hspace{0.32em}}
\newcommand{\ckdot}[1]{\tikz[baseline=-0.05ex]\fill[#1] (0.31em,0.31em) circle (0.29em);\hspace{0.32em}}
\newcommand{\ckdia}[1]{\tikz[baseline=-0.05ex]\fill[#1]
  (0.31em,-0.04em) -- (0.66em,0.31em) -- (0.31em,0.66em) -- (-0.04em,0.31em) -- cycle;\hspace{0.32em}}
\newcommand{\ckmodel}[2]{\modellogo{#1}#2}
\newcommand{\ckharness}[2]{\harnesslogo{#1}#2}

\renewcommand{\FigTaxonomy}{%
\resizebox{\linewidth}{!}{%
\begin{tikzpicture}[x=1cm,y=1cm]
  \useasboundingbox (0,-4.55) rectangle (13.4,0.2);
  \node[ckTitle,anchor=west] at (0,0) {(a) Category breakdown};
  \node[ckSub,anchor=west] at (0,-0.37) {Failures and induced groups in each fixed category, count};
  \def\cF{2.70}\def\cG{5.25}\def\s{0.0075}
  \node[ckHead,anchor=west] at (\cF,-0.80) {Failures};
  \node[ckHead,anchor=west] at (\cG,-0.80) {Induced groups};
  \draw[ckHeadRule] (0,-0.98) -- (7.55,-0.98);
  \foreach \i/\nm/\grp/\cnt in {
      0/{Incomplete work}/163/264,1/{Instruction violation}/107/179,
      2/{Verification failure}/102/165,3/{Tool misuse}/51/100,
      4/{Reasoning error}/47/88,5/{Planning}/45/68,
      6/{Hallucination}/44/67,7/{Environment error}/17/25}{
    \pgfmathsetmacro{\y}{-1.25-\i*0.42}
    \node[ckLabel,anchor=west] at (0,\y) {\nm};
    \fill[ckGrey!70] (\cF,{\y-0.125}) rectangle ({\cF+\cnt*\s},{\y+0.125});
    \node[ckNum,anchor=west] at ({\cF+\cnt*\s+0.08},\y) {\cnt};
    \fill[ckMain] (\cG,{\y-0.125}) rectangle ({\cG+\grp*\s},{\y+0.125});
    \node[ckValue,text=ckMain,anchor=west] at ({\cG+\grp*\s+0.08},\y) {\grp};}
  \draw[ckEndRule] (0,-4.43) -- (7.55,-4.43);
  \node[ckTitle,anchor=west] at (8.25,0) {(b) Group granularity};
  \node[ckSub,anchor=west] at (8.25,-0.37) {Summary statistics of the two representations};
  \fill[ckBand] (12.05,-0.58) rectangle (13.4,-4.43);
  \node[ckHead,anchor=east] at (11.80,-0.80) {Fixed};
  \node[ckHead,text=ckMain,anchor=east] at (13.25,-0.80) {Induced};
  \draw[ckHeadRule] (8.25,-0.98) -- (13.4,-0.98);
  \foreach \i/\nm/\a/\b in {
      0/{Groups}/8/505,
      1/{Largest group}/264/11,
      2/{Largest share}/{27.5\%}/{1.1\%},
      3/{Median size}/94/1,
      4/{Singleton groups}/0/381,
      5/{Unassigned}/5/0}{
    \pgfmathsetmacro{\y}{-1.30-\i*0.57}
    \node[ckLabel,anchor=west] at (8.25,\y) {\nm};
    \ifnum\i=0
      \node[ckNum,anchor=east,font=\sffamily\fontsize{9.6}{11}\selectfont] at (11.80,\y) {\a};
      \node[ckValue,text=ckMain,anchor=east,font=\sffamily\bfseries\fontsize{9.6}{11}\selectfont]
        at (13.25,\y) {\b};
    \else
      \node[ckNum,anchor=east] at (11.80,\y) {\a};
      \node[ckValue,text=ckMain,anchor=east] at (13.25,\y) {\b};
    \fi
    \ifnum\i<5\draw[ckGrid] (8.25,{\y-0.285}) -- (13.4,{\y-0.285});\fi}
  \draw[ckEndRule] (8.25,-4.43) -- (13.4,-4.43);
\end{tikzpicture}}}

\renewcommand{\FigPairShare}{%
\resizebox{\linewidth}{!}{%
\begin{tikzpicture}[x=1cm,y=1cm]
  \useasboundingbox (0,-2.05) rectangle (13.4,0.2);
  \node[ckTitle,anchor=west] at (0,0) {Pairs that share at least one induced group};
  \node[ckSub,anchor=west] at (0,-0.37) {Share of configuration pairs, \%};
  \def\L{3.55}\def\W{6.70}
  \foreach \y/\nm/\desc/\p/\n/\d/\col/\vc in {
      -0.98/{Related pairs}/{Model version or harness changed}/44.4/4/9/ckMain/ckMain,
      -1.68/{Other pairs}/{All remaining configuration pairs}/14.4/26/181/ckGrey/ckText}{
    \node[ckLabel,text=ckInk,font=\sffamily\bfseries\fontsize{7.4}{9}\selectfont,anchor=west]
      at (0,{\y+0.11}) {\nm};
    \node[ckNote,anchor=west] at (0,{\y-0.19}) {\desc};
    \fill[ckTrack] (\L,{\y-0.17}) rectangle ({\L+\W},{\y+0.17});
    \fill[\col] (\L,{\y-0.17}) rectangle ({\L+\p/100*\W},{\y+0.17});
    \node[ckValue,text=\vc,anchor=west] at ({\L+\p/100*\W+0.10},\y)
      {\p\%\enspace{\mdseries\fontsize{6.6}{8}\selectfont\color{ckMuted}\n\ of \d\ pairs}};}
  \draw[ckInk,line width=0.5pt] ({\L+\W+0.22},-0.98) -- ({\L+\W+0.36},-0.98)
    -- ({\L+\W+0.36},-1.68) -- ({\L+\W+0.22},-1.68);
  \node[anchor=west,inner sep=0pt,text=ckAccent,font=\sffamily\bfseries\fontsize{11}{12}\selectfont]
    at ({\L+\W+0.52},-1.21) {3.1×};
  \node[ckNote,anchor=west] at ({\L+\W+0.52},-1.52) {higher for related pairs};
\end{tikzpicture}}}

\newcommand{\FigRecurrenceCost}{%
\resizebox{\linewidth}{!}{%
\begin{tikzpicture}[x=1cm,y=1cm]
  \useasboundingbox (0,-5.95) rectangle (13.4,0.2);
  \node[ckTitle,anchor=west] at (0,0) {(a) Failure recurrence};
  \node[ckSub,anchor=west] at (0,-0.37) {Proposed failure explanations, count};
  \def\L{2.45}\def\s{0.0119}
  \foreach \y/\nm/\a/\b/\col/\lab/\tc in {
      -0.85/{Proposed}/0/239/ckMain/{239}/ckInk,
      -1.27/{Not tested in budget}/119/239/ckLight/{−120}/ckMuted,
      -1.69/{Tested on siblings}/0/119/ckMain/{119}/ckInk}{
    \node[ckLabel,anchor=west] at (0,\y) {\nm};
    \fill[\col] ({\L+\a*\s},{\y-0.13}) rectangle ({\L+\b*\s},{\y+0.13});
    \node[ckValue,text=\tc,anchor=west] at ({\L+\b*\s+0.08},\y) {\lab};}
  \draw[ckStep] ({\L+239*\s},-0.98) -- ({\L+239*\s},-1.14);
  \draw[ckStep] ({\L+119*\s},-1.40) -- ({\L+119*\s},-1.56);
  \node[ckNote,anchor=west] at (0,-2.07) {Outcome of the 119 tested explanations};
  \draw[ckGrid] (0,-1.92) -- (6.30,-1.92);
  \foreach \y/\nm/\a/\n/\pct/\col/\tc in {
      -2.43/{Agent behavior}/0/20/{16.8\%}/ckAccent/ckAccent,
      -2.85/{Weak recurrence}/20/26/{21.8\%}/ckAccentSoft/ckInk,
      -3.27/{Task-specific}/46/48/{40.3\%}/ckGrey/ckInk,
      -3.69/{Unverified}/94/25/{21.0\%}/ckLight/ckInk}{
    \node[ckLabel,anchor=west] at (0,\y) {\nm};
    \fill[\col] ({\L+\a*\s},{\y-0.13}) rectangle ({\L+(\a+\n)*\s},{\y+0.13});
    \node[ckValue,text=\tc,anchor=west] at ({\L+(\a+\n)*\s+0.08},\y) {\n};
    \node[ckNote,anchor=west] at ({\L+(\a+\n)*\s+0.40},\y) {\pct};}
  \foreach \cx/\ya/\yb in {20/-2.56/-2.72,46/-2.98/-3.14,94/-3.40/-3.56}{
    \draw[ckStep] ({\L+\cx*\s},\ya) -- ({\L+\cx*\s},\yb);}
  \node[ckNote,anchor=west] at (0,-4.00) {Unverified: fewer than three tested siblings.};
  \node[ckTitle,anchor=west] at (6.95,0) {(b) Replay cost and failure rate};
  \node[ckSub,anchor=west] at (6.95,-0.37) {Matched failure rate by input tokens per rollout};
  \def\PL{7.75}\def\PW{5.50}\def\PB{-3.42}\def\PH{2.58}
  \def\XA{5.0}\def\XB{6.65}
  \foreach \v/\lab in {5/{100k},5.30103/{200k},5.69897/{500k},6/{1M},6.30103/{2M},6.60206/{4M}}{
    \pgfmathsetmacro{\x}{\PL+(\v-\XA)/(\XB-\XA)*\PW}
    \node[ckTick,anchor=north] at (\x,{\PB-0.10}) {\lab};}
  \foreach \v/\lab in {0/{0\%},50/{50\%},100/{100\%}}{
    \pgfmathsetmacro{\y}{\PB+\v/100*\PH}
    \ifnum\v>0\draw[ckGrid] (\PL,\y) -- ({\PL+\PW},\y);\fi
    \node[ckTick,anchor=east] at ({\PL-0.10},\y) {\lab};}
  \draw[ckBase] (\PL,\PB) -- ({\PL+\PW},\PB);
  \node[ckNote,anchor=north] at ({\PL+\PW/2},{\PB-0.40}) {Input tokens per rollout, log scale};
  \foreach \id/\lx/\rate/\dx/\dy in {
      1/5.0479/66.7/.17/.16,2/5.0732/22.2/.17/.13,
      3/5.1173/38.9/.17/-.15,4/5.1665/55.6/.17/.14,
      5/5.3890/50.0/-.15/.17,6/5.5167/72.2/-.15/.18,
      7/5.5308/33.3/-.15/-.17,8/5.5626/44.4/.17/-.14,
      9/5.6215/66.7/0/.19,10/5.6307/66.7/0/.19,
      11/5.6730/50.0/.18/.14,12/5.9235/94.4/.18/.10,
      13/6.0519/27.8/.20/-.15,14/6.0620/72.2/.18/.14,
      15/6.5427/77.8/-.22/.14}{
    \pgfmathsetmacro{\x}{\PL+(\lx-\XA)/(\XB-\XA)*\PW}
    \pgfmathsetmacro{\y}{\PB+\rate/100*\PH}
    \ifnum\id=13
      \fill[white] (\x,{\y+.12}) -- ({\x+.12},\y) -- (\x,{\y-.12}) -- ({\x-.12},\y) -- cycle;
      \fill[ckAccent] (\x,{\y+.085}) -- ({\x+.085},\y) -- (\x,{\y-.085}) -- ({\x-.085},\y) -- cycle;
    \else\ifnum\id=15
      \fill[white] (\x,{\y+.12}) -- ({\x+.12},\y) -- (\x,{\y-.12}) -- ({\x-.12},\y) -- cycle;
      \fill[ckAccent] (\x,{\y+.085}) -- ({\x+.085},\y) -- (\x,{\y-.085}) -- ({\x-.085},\y) -- cycle;
    \else
      \fill[white] (\x,\y) circle (0.085);
      \fill[ckMain] (\x,\y) circle (0.058);
    \fi\fi
    \ifnum\id=9
      \node[ckTick,text=ckText] at ({\x+.02},{\y+.21}) {9,10};
    \else\ifnum\id=10\else
      \node[ckTick,text=ckText] at ({\x+\dx},{\y+\dy}) {\id};
    \fi\fi}
  \draw[ckHeadRule] (0,-4.33) -- (13.4,-4.33);
  \node[ckHead,anchor=west] at (0,-4.56) {Configurations in (b)};
  \node[ckNote,anchor=west] at (2.85,-4.56)
    {\ckdot{ckMain}Terminus 2\quad\ckdia{ckAccent}Codex or Claude Code};
  \node[ckNote,anchor=east] at (13.4,-4.56)
    {Terminus 2: Spearman {\itshape ρ} = 0.50, {\itshape p} = 0.08, {\itshape n} = 13};
  \foreach \id/\name/\col/\vendor [count=\j from 0] in {
      1/{Claude Sonnet 5}/ckMain/cl,2/{Kimi K3}/ckMain/ms,3/{GPT-5.6-sol}/ckMain/oa,
      4/{Qwen3.8-Max}/ckMain/qw,5/{GLM-5.2}/ckMain/za,6/{Grok 4.5}/ckMain/xa,
      7/{GLM-5}/ckMain/za,8/{Claude Opus 5}/ckMain/cl,9/{MiniMax-M2.7}/ckMain/mm,
      10/{DeepSeek-V4}/ckMain/ds,11/{Claude Sonnet 4.5}/ckMain/cl,12/{GPT-5.4-nano}/ckMain/oa,
      13/{GPT-5.6-sol\enspace\harnesslogo{cx}Codex}/ckAccent/oa,14/{Claude Opus 4.8}/ckMain/cl,
      15/{Sonnet 4.5\enspace\harnesslogo{cc}Claude Code}/ckAccent/cl}{
    \pgfmathtruncatemacro{\cc}{floor(\j/4)}
    \pgfmathtruncatemacro{\rr}{mod(\j,4)}
    \pgfmathsetmacro{\xx}{\cc*3.38}
    \ifnum\cc=3\def\xx{9.80}\fi
    \pgfmathsetmacro{\yy}{-4.92-\rr*0.30}
    \node[ckTick,font=\sffamily\bfseries\fontsize{6.8}{8}\selectfont,text=\col,anchor=east]
      at ({\xx+0.25},\yy) {\id};
    \node[ckTick,text=ckText,font=\sffamily\fontsize{6.9}{8}\selectfont,anchor=west]
      at ({\xx+0.36},\yy) {\ckmodel{\vendor}{\name}};}
\end{tikzpicture}}}

\newcommand{\FigAblation}{%
\resizebox{\linewidth}{!}{%
\begin{tikzpicture}[x=1cm,y=1cm]
  \useasboundingbox (0,-3.95) rectangle (13.4,0.2);
  \node[ckTitle,anchor=west] at (0,0) {Induction ablations};
  \node[ckSub,anchor=west] at (0,-0.37) {961 failures, similarity margin {\itshape μ} = 0.05};
  \def\AL{3.45}\def\AW{3.0}
  \def\CL{8.60}\def\CS{0.00015}
  \node[ckHead,anchor=west] at (0,-0.80) {Variant};
  \node[ckHead,anchor=west] at (\AL,-0.80) {Held-out agreement};
  \node[ckHead,anchor=east] at (8.10,-0.80) {Δ vs.\ full};
  \node[ckHead,anchor=west] at (\CL,-0.80) {Judge calls};
  \node[ckHead,anchor=east] at (12.40,-0.80) {Groups};
  \node[ckHead,anchor=east] at (13.40,-0.80) {Shared};
  \fill[ckBand] (0,-1.01) rectangle (13.4,-1.48);
  \draw[ckHeadRule] (0,-0.99) -- (13.4,-0.99);
  \foreach \i/\name/\g/\shared/\calls/\callslab/\acc/\dlt/\acol/\dcol in {
      0/{Full method}/505/11/7449/{7,449}/91.8/{--}/ckMain/ckMuted,
      1/{No density constraint}/23/4/7449/{7,449}/78.0/{−13.8}/ckAccent/ckAccent,
      2/{Fixed thresholds}/500/11/6789/{6,789}/90.8/{−1.0}/ckGrey/ckText,
      3/{Neighbors {\itshape k}\textsubscript{nn} = 5}/506/10/2614/{2,614}/93.3/{+1.5}/ckGrey/ckText,
      4/{Neighbors {\itshape k}\textsubscript{nn} = 30}/505/11/14297/{14,297}/92.4/{+0.6}/ckGrey/ckText,
      5/{Single threshold}/500/11/0/{0}/92.1/{+0.3}/ckGrey/ckText}{
    \pgfmathsetmacro{\y}{-1.245-\i*0.47}
    \ifnum\i=0
      \node[ckLabel,text=ckInk,font=\sffamily\bfseries\fontsize{7.4}{9}\selectfont,anchor=west]
        at (0,\y) {\name};
    \else
      \node[ckLabel,anchor=west] at (0,\y) {\name};
      \draw[ckGrid] (0,{\y+0.235}) -- (13.4,{\y+0.235});
    \fi
    \fill[\acol] (\AL,{\y-0.12}) rectangle ({\AL+\acc/100*\AW},{\y+0.12});
    \ifnum\i<2\def\vcol{\acol}\else\def\vcol{ckText}\fi
    \node[ckValue,text=\vcol,anchor=west] at ({\AL+\acc/100*\AW+0.08},\y) {\acc\%};
    \ifnum\i=1
      \node[ckValue,text=ckAccent,anchor=east] at (8.10,\y) {\dlt};
    \else
      \node[ckNum,text=\dcol,anchor=east] at (8.10,\y) {\dlt};
    \fi
    \ifnum\i=0\def\ccol{ckMain}\else\def\ccol{ckGrey}\fi
    \ifnum\calls>0
      \fill[\ccol] (\CL,{\y-0.12}) rectangle ({\CL+\calls*\CS},{\y+0.12});
    \fi
    \node[ckNum,anchor=west] at ({\CL+\calls*\CS+0.08},\y) {\callslab};
    \node[ckNum,anchor=east] at (12.40,\y) {\g};
    \node[ckNum,anchor=east] at (13.40,\y) {\shared};}
  \draw[ckEndRule] (0,-3.81) -- (13.4,-3.81);
\end{tikzpicture}}}

\renewcommand{\FigGranularity}{%
\resizebox{\linewidth}{!}{%
\begin{tikzpicture}[x=1cm,y=1cm]
  \useasboundingbox (0,-4.42) rectangle (13.4,0.2);
  \def\SL{0.70}\def\B{-3.25}\def\HH{2.30}
  \node[ckTitle,anchor=west] at (0,0) {(a) Group counts};
  \node[ckSub,anchor=west] at (0,-0.37)
    {\cksq{ckMain}Two or more failures\quad\cksq{ckLight}Single failure};
  \def\L{1.30}
  \foreach \i/\lab/\tot/\multi in {0/None/23/17,1/0/374/168,2/.02/421/155,
      3/.05/505/124,4/.08/565/101,5/.12/603/76,6/.20/665/38}{
    \pgfmathsetmacro{\x}{\L+0.35+\i*\SL}
    \fill[ckMain] ({\x-0.21},\B) rectangle ({\x+0.21},{\B+\multi/700*\HH});
    \fill[ckLight] ({\x-0.21},{\B+\multi/700*\HH}) rectangle ({\x+0.21},{\B+\tot/700*\HH});
    \node[ckValue,anchor=south] at (\x,{\B+\tot/700*\HH+0.06}) {\tot};
    \node[ckTick,text=ckText,anchor=north] at (\x,{\B-0.12}) {\lab};
    \node[ckValue,text=ckMain,anchor=north] at (\x,{\B-0.47}) {\multi};}
  \draw[ckBase] (\L,\B) -- ({\L+7*\SL},\B);
  \node[ckNote,anchor=west] at (0,{\B-0.20}) {Margin {\itshape μ}};
  \node[ckNote,anchor=west] at (0,{\B-0.55}) {Two or more};
  \node[ckNote,anchor=west] at (0,{\B-0.98}) {No constraint: the largest group contains 650 failures.};
  \node[ckTitle,anchor=west] at (7.0,0) {(b) Description agreement};
  \node[ckSub,anchor=west] at (7.0,-0.37) {Held-out agreement on 961 failures, \%};
  \def\R{8.30}
  \draw[ckBase] (\R,\B) -- ({\R+7*\SL},\B);
  \draw[ckLink] ({\R+0.35},{\B+(70.7-65)/31*\HH}) -- ({\R+0.35+\SL},{\B+(87.4-65)/31*\HH});
  \draw[ckMain,line width=1.1pt]
    \foreach \i/\acc [count=\j] in {1/87.4,2/90.2,3/91.4,4/92.1,5/92.6,6/93.1}{
      \ifnum\j>1 -- \fi ({\R+0.35+\i*\SL},{\B+(\acc-65)/31*\HH})};
  \foreach \i/\lab/\acc in {0/None/70.7,1/0/87.4,2/.02/90.2,3/.05/91.4,4/.08/92.1,5/.12/92.6,6/.20/93.1}{
    \pgfmathsetmacro{\x}{\R+0.35+\i*\SL}
    \pgfmathsetmacro{\y}{\B+(\acc-65)/31*\HH}
    \fill[white] (\x,\y) circle (0.09);
    \ifnum\i=0
      \fill[ckAccent] (\x,\y) circle (0.062);
      \node[ckValue,text=ckAccent,anchor=south] at (\x,{\y+0.11}) {\acc};
    \else
      \fill[ckMain] (\x,\y) circle (0.062);
      \node[ckValue,text=ckMain,anchor=south] at (\x,{\y+0.11}) {\acc};
    \fi
    \node[ckTick,text=ckText,anchor=north] at (\x,{\B-0.12}) {\lab};}
  \node[ckNote,anchor=west] at (7.0,{\B-0.20}) {Margin {\itshape μ}};
\end{tikzpicture}}}

\renewcommand{\FigCalibration}{%
\resizebox{\linewidth}{!}{%
\begin{tikzpicture}[x=1cm,y=1cm]
  \useasboundingbox (0,-2.30) rectangle (13.4,0.2);
  \node[ckTitle,anchor=west] at (0,0) {(c) Threshold calibration};
  \node[ckSub,anchor=west] at (0,-0.37) {Similarity band sent to the pairwise judge};
  \def\L{2.85}\def\W{6.80}
  \def\RL{10.50}\def\RW{2.10}
  \node[ckHead,anchor=west] at (\L,-0.80) {Centered cosine similarity};
  \node[ckHead,anchor=west] at (\RL,-0.80) {Pairs labeled matching};
  \draw[ckHeadRule] (0,-0.98) -- (13.4,-0.98);
  \draw[ckRule,line width=0.45pt] ({\L+\W/2},-1.05) -- ({\L+\W/2},-2.00);
  \foreach \y/\lab/\range/\a/\b/\pct/\col/\tc in {
      -1.30/{Fixed thresholds}/{[−0.10, 0.15]}/-0.10/0.15/9/ckGrey/ckText,
      -1.80/{Calibrated thresholds}/{[−0.03, 0.64]}/-0.03/0.64/41/ckMain/ckMain}{
    \node[ckLabel,anchor=west] at (0,\y) {\lab};
    \fill[ckTrack] (\L,{\y-0.13}) rectangle ({\L+\W},{\y+0.13});
    \fill[\col] ({\L+(\a+1)/2*\W},{\y-0.13}) rectangle ({\L+(\b+1)/2*\W},{\y+0.13});
    \node[ckNote,text=ckText,anchor=west] at ({\L+(\b+1)/2*\W+0.10},\y) {\range};
    \fill[\col] (\RL,{\y-0.13}) rectangle ({\RL+\pct/100*\RW},{\y+0.13});
    \node[ckValue,text=\tc,anchor=west] at ({\RL+\pct/100*\RW+0.08},\y) {\pct\%};}
  \foreach \v/\lab in {-1/{−1},-0.5/{−0.5},0/0,0.5/0.5,1/1}{
    \node[ckTick,anchor=north] at ({\L+(\v+1)/2*\W},-2.05) {\lab};}
\end{tikzpicture}}}

\renewcommand{\FigReplay}{%
\resizebox{\linewidth}{!}{%
\begin{tikzpicture}[x=1cm,y=1cm]
  \useasboundingbox (0,-7.05) rectangle (13.4,0.2);
  \def\L{5.65}\def\W{7.55}
  \node[ckHead,anchor=west] at (0,0) {Model};
  \node[ckHead,anchor=west] at (3.55,0) {Harness};
  \node[ckHead,anchor=west] at (\L,0) {Matched failure rate, \%};
  \draw[ckHeadRule] (0,-0.20) -- (13.4,-0.20);
  \foreach \i/\est/\pct/\lo/\hi/\model/\harness/\other/\vendor/\hk in {
      0/.222/{22.2}/.090/.452/{Kimi K3}/{Terminus 2}/0/ms/t2,
      1/.278/{27.8}/.125/.509/{GPT-5.6-sol}/{Codex}/1/oa/cx,
      2/.333/{33.3}/.163/.563/{GLM-5}/{Terminus 2}/0/za/t2,
      3/.389/{38.9}/.203/.614/{GPT-5.6-sol}/{Terminus 2}/0/oa/t2,
      4/.444/{44.4}/.246/.663/{Claude Opus 5}/{Terminus 2}/0/cl/t2,
      5/.500/{50.0}/.290/.710/{GLM-5.2}/{Terminus 2}/0/za/t2,
      6/.500/{50.0}/.290/.710/{Claude Sonnet 4.5}/{Terminus 2}/0/cl/t2,
      7/.556/{55.6}/.337/.754/{Qwen3.8-Max}/{Terminus 2}/0/qw/t2,
      8/.667/{66.7}/.437/.837/{DeepSeek-V4}/{Terminus 2}/0/ds/t2,
      9/.667/{66.7}/.437/.837/{MiniMax-M2.7}/{Terminus 2}/0/mm/t2,
      10/.667/{66.7}/.437/.837/{Claude Sonnet 5}/{Terminus 2}/0/cl/t2,
      11/.722/{72.2}/.491/.875/{Grok 4.5}/{Terminus 2}/0/xa/t2,
      12/.722/{72.2}/.491/.875/{Claude Opus 4.8}/{Terminus 2}/0/cl/t2,
      13/.778/{77.8}/.548/.910/{Claude Sonnet 4.5}/{Claude Code}/1/cl/cc,
      14/.944/{94.4}/.742/.990/{GPT-5.4-nano}/{Terminus 2}/0/oa/t2}{
    \pgfmathsetmacro{\y}{-0.48-\i*0.40}
    \ifnum\other=1\def\c{ckAccent}\else\def\c{ckMain}\fi
    \node[ckLabel,anchor=west] at (0,\y) {\ckmodel{\vendor}{\model}};
    \node[ckLabel,text=ckMuted,anchor=west] at (3.55,\y) {\ckharness{\hk}{\harness}};
    \fill[ckTrack] (\L,{\y-0.14}) rectangle ({\L+\W},{\y+0.14});
    \fill[\c!22] ({\L+\lo*\W},{\y-0.14}) rectangle ({\L+\hi*\W},{\y+0.14});
    \node[ckValue,text=white,fill=\c,inner xsep=2.6pt,inner ysep=1.4pt,rounded corners=1.2pt]
      at ({\L+\est*\W},\y) {\pct};}
  \foreach \v/\lab in {0/{0\%},0.25/{25\%},0.5/{50\%},0.75/{75\%},1/{100\%}}{
    \node[ckTick,anchor=north] at ({\L+\v*\W},-6.24) {\lab};}
  \node[ckNote,anchor=west] at (0,-6.80)
    {\cksq{ckMain}Terminus 2\quad\cksq{ckAccent}Codex or Claude Code};
  \node[ckNote,anchor=east] at (13.4,-6.80)
    {Label: rate over 18 matched trials\quad Tinted band: 95\% Wilson interval};
\end{tikzpicture}}}

\renewcommand{\FigPairs}{%
\resizebox{\linewidth}{!}{%
\begin{tikzpicture}[x=1cm,y=1cm]
  \useasboundingbox (0,-3.55) rectangle (13.4,0.2);
  \def\s{0.0330}
  \foreach \px/\bo/\title/\sub in {0/2.85/{(a) Model-version pairs}/{Distinct failure groups; Terminus 2 held fixed},
      7.05/3.42/{(b) Harness pairs}/{Distinct failure groups; same model}}{
    \node[ckTitle,anchor=west] at (\px,0) {\title};
    \node[ckSub,anchor=west] at (\px,-0.37) {\sub};
    \node[ckHead,anchor=west] at ({\px+\bo},-0.80) {Groups};
    \node[ckHead,anchor=east] at ({\px+6.35},-0.80) {Change};
    \draw[ckHeadRule] (\px,-0.98) -- ({\px+6.35},-0.98);
    \foreach \v in {0,35,70}{
      \node[ckTick,anchor=north] at ({\px+\bo+\v*\s},-3.02) {\v};}}
  \foreach \px/\bo/\i/\two/\close/\name/\note/\va/\vb/\dl/\vendor in {
      0/2.85/0/0/1/{Sonnet 4.5 $\to$ 5}/{}/52/56/{+4}/cl,
      0/2.85/1/0/0/{GLM-5 $\to$ 5.2}/{}/35/6/{−29}/za,
      0/2.85/2/0/0/{Opus 4.8 $\to$ 5}/{}/8/24/{+16}/cl,
      7.05/3.42/0/1/0/{Sonnet 4.5}/{\harnesslogo{cc}Claude Code $\to$ \harnesslogo{t2}Terminus 2}/16/52/{+36}/cl,
      7.05/3.42/1/1/0/{GPT-5.6-sol}/{\harnesslogo{cx}Codex $\to$ \harnesslogo{t2}Terminus 2}/7/58/{+51}/oa}{
    \pgfmathsetmacro{\y}{-1.33-\i*0.62}
    \ifnum\i>0\draw[ckGrid] (\px,{\y+0.31}) -- ({\px+6.35},{\y+0.31});\fi
    \ifnum\two=0
      \node[ckLabel,anchor=west] at (\px,\y) {\ckmodel{\vendor}{\name}};
    \else
      \node[ckLabel,anchor=west] at (\px,{\y+0.11}) {\ckmodel{\vendor}{\name}};
      \node[ckNote,anchor=west] at (\px,{\y-0.17}) {\note};
    \fi
    \pgfmathsetmacro{\xa}{\px+\bo+\va*\s}
    \pgfmathsetmacro{\xb}{\px+\bo+\vb*\s}
    \fill[ckTrack] ({\px+\bo},{\y-0.14}) rectangle ({\px+\bo+70*\s},{\y+0.14});
    \fill[ckMain!18] (\xa,{\y-0.14}) rectangle (\xb,{\y+0.14});
    \ifnum\close=1
      \node[ckValue,text=white,fill=ckGrey,inner xsep=2.6pt,inner ysep=1.4pt,rounded corners=1.2pt,
        anchor=east] at (\xa,\y) {\va};
      \node[ckValue,text=white,fill=ckMain,inner xsep=2.6pt,inner ysep=1.4pt,rounded corners=1.2pt,
        anchor=west] at (\xb,\y) {\vb};
    \else
      \node[ckValue,text=white,fill=ckGrey,inner xsep=2.6pt,inner ysep=1.4pt,rounded corners=1.2pt]
        at (\xa,\y) {\va};
      \node[ckValue,text=white,fill=ckMain,inner xsep=2.6pt,inner ysep=1.4pt,rounded corners=1.2pt]
        at (\xb,\y) {\vb};
    \fi
    \node[ckValue,anchor=east] at ({\px+6.35},\y) {\dl};}
  \node[ckNote,anchor=west] at (0,-3.42)
    {\cksq{ckGrey}First configuration\qquad\cksq{ckMain}Second configuration\qquad
     Tinted band: change between them};
\end{tikzpicture}}}

\newcommand{\probeagent}{\textsc{ProbeAgent}}
\newcommand{\TODO}[1]{}

\title{\textsc{ProbeAgent}: Principled Diagnosis of Agent Failures in Verifiable Environments}
\newcommand{\preprintdisplaytitle}{\textcolor{ckAccent}{ProbeAgent}: Principled Diagnosis of Agent Failures in Verifiable Environments}
\author[1]{Yue~Huang}
\author[2]{Yuchen~Ma}
\author[3]{Shuhang~Lin}
\author[4]{Yuyang~Wu}
\author[5]{Zixuan~Weng}
\author[6]{Zixiang~Xu}
\author[1]{Xiangqi~Wang}
\author[7]{Yuexing~Hao}
\author[1]{Yujun~Zhou}
\author[6]{Yue~Zhao}
\author[4]{Olexandr~Isayev}
\author[2]{Stefan~Feuerriegel}
\author[1]{Xiangliang~Zhang}

\affil[1]{University of Notre Dame}
\affil[2]{Ludwig-Maximilians-Universität München}
\affil[3]{Rutgers University}
\affil[4]{Carnegie Mellon University}
\affil[5]{Georgia Institute of Technology}
\affil[6]{University of Southern California}
\affil[7]{Massachusetts Institute of Technology}

\newcommand{\preprintauthor}[2]{#1\textsuperscript{#2}}
\newcommand{\preprintaffil}[3]{\textsuperscript{#1}#3}
\newcommand{\preprintauthorlist}[1]{%
  \preprintauthor{Yue Huang}{1}#1\preprintauthor{Yuchen Ma}{2}#1%
  \preprintauthor{Shuhang Lin}{3}#1\preprintauthor{Yuyang Wu}{4}#1%
  \preprintauthor{Zixuan Weng}{5}#1\preprintauthor{Zixiang Xu}{6}#1%
  \preprintauthor{Xiangqi Wang}{1}#1\preprintauthor{Yuexing Hao}{7}#1%
  \preprintauthor{Yujun Zhou}{1}#1\preprintauthor{Yue Zhao}{6}#1%
  \preprintauthor{Olexandr Isayev}{4}#1\preprintauthor{Stefan Feuerriegel}{2}#1%
  \preprintauthor{Xiangliang Zhang}{1}}
\newcommand{\preprintaffillist}[1]{%
  \preprintaffil{1}{nd}{University of Notre Dame}#1%
  \preprintaffil{2}{lmu}{Ludwig-Maximilians-Universität München}#1%
  \preprintaffil{3}{ru}{Rutgers University}#1%
  \preprintaffil{4}{cmu}{Carnegie Mellon University}#1%
  \preprintaffil{5}{gt}{Georgia Institute of Technology}#1%
  \preprintaffil{6}{usc}{University of Southern California}#1%
  \preprintaffil{7}{mit}{Massachusetts Institute of Technology}}
\hypersetup{pdfauthor={Yue Huang, Yuchen Ma, Shuhang Lin, Yuyang Wu, Zixuan Weng,
  Zixiang Xu, Xiangqi Wang, Yuexing Hao, Yujun Zhou, Yue Zhao, Olexandr Isayev,
  Stefan Feuerriegel, Xiangliang Zhang}}

\begin{abstract}
Improving agents requires discovering where they fail and understanding why.
An agent acts over multiple steps in a changing environment, so diagnosing its
failures requires more than inspecting the final response of a large language
model (LLM). The same unsuccessful outcome can arise from task difficulty or from
an error the agent repeats across tasks. We introduce \probeagent{}, an automated
framework that discovers agent failures, tests their proposed causes, and identifies
shared failure modes. Starting from benchmark tasks, \probeagent{} builds executable
tests and validates them using reference solutions, automated checks, and review
of the task requirements. For each failed trajectory, it explains the error and
constructs new tasks that preserve the suspected cause while changing the surrounding
context. Running the agent on these tasks tests whether the same error recurs.
\probeagent{} then groups failures with similar causes and generates descriptions
that distinguish their errors. This process connects explanations of individual
failures to executable tests and descriptions of shared failure modes.
We evaluate \probeagent{} across 22 agent configurations using tasks from 10 benchmarks,
collecting 1,244 failures and producing 536 failure-mode descriptions. These descriptions
distinguish specific errors within broad outcomes such as incomplete work, including
planning without acting and stopping before delivering a file. In human evaluation,
annotators accept 90.0\% of assigned descriptions and correctly reject 83.3\% of similar
alternatives, supporting the accuracy and specificity of the discovered failure modes.

\end{abstract}
\begin{document}
\makepreprinttitle
\section{Introduction}

Agents are increasingly used to modify code, navigate websites, and complete work in
terminal environments \citep{jimenez2024swebench,zhou2024webarena,terminalbench2026}.
An agent combines a large language model (LLM) with a \emph{harness}, which manages
its context, execution flow, and termination. As agents become part of real-world
workflows, they still fail on many tasks. Studying these failures can guide improvement:
we need to identify where agents fail and understand why. Benchmark scores summarize
task success, but the final outcome alone does not explain the cause of a failure.
The same outcome can arise for different reasons. For example, a missing report may
result from a difficult computation, or simply because the agent stops before saving
its result. These failures require different fixes. Improving agents therefore requires
not only measuring failures, but also systematically discovering them and diagnosing
their causes.

Existing automated evaluation methods often discover LLM failures by generating
new questions or selecting informative tests
\citep{cheng2024autodetect,li2025autobencher,afkanpour2025ace}.
ProbeLLM \citep{huang2026probellm} goes beyond finding individual errors by explaining
incorrect responses and organizing related errors into \emph{failure modes}.
Each mode describes a shared error pattern supported by concrete failure examples.
This connects the discovery of individual failures with an account of the patterns
shared across them.

For agents, existing methods diagnose failed trajectories after execution.
For example, Who\&When benchmarks failure attribution \citep{zhang2025whowhen},
while AgentRx and AgentDebug analyze
\emph{trajectories}, the sequences of actions and observations, to locate errors or
guide corrective feedback \citep{barke2026agentrx,zhu2025agentdebug}.
These studies help explain observed failures, but leave open how to systematically
construct new tasks to discover further failures and organize them into shared modes.
We therefore ask: \emph{can we build an effective framework that automatically discovers
and diagnoses agent failures?}

This requires addressing three challenges.
\textbf{(1) Building reliable executable tests.}
An LLM test can pair a question with a reference answer and a verifier. An agent test
also needs a working environment in which actions change the task state, a solvable
task, and a checker that evaluates the required outcome. Environment defects or
shortcuts accepted by the checker can otherwise give misleading results
\citep{aleithan2024swebenchplus,wang2026benchjack}.
\textbf{(2) Distinguishing task-specific difficulties from recurring agent behavior.}
A failed trajectory may reflect a difficulty in one task or a behavior that appears in many
tasks. Explaining that trajectory alone does not distinguish the two. New tests must change
the task context while preserving the condition thought to cause the failure, so that
we can check whether the same agent makes the same mistake again.
\textbf{(3) Identifying shared failures across different tasks.}
Omitting a CSV export after data analysis and a JSON export after a database query
can reflect the same error: completing the computation without delivering its output.
A failed write can also leave a file missing, but requires a different correction.
Diagnosis must distinguish the behaviors behind these similar outcomes.

To address these challenges, we propose \probeagent{}, an automated framework for
 discovering and diagnosing agent failures\ifprobeagentpreprint{} (Figure~\ref{fig:intro-overview})\fi. It connects three stages.
\emph{Establish failures} constructs executable probes from benchmark tasks,
validates their checkers and reference solutions, and explains failed trajectories.
\emph{Test failure recurrence} constructs \emph{sibling tasks}: new tasks that preserve
the proposed triggering condition while changing the domain, data, or output format.
Running the same agent on these tasks tests whether the explained error recurs.
\emph{Discover failure modes} groups failures by their causes and generates descriptions
that distinguish their errors. By using explanations to guide new executable tests,
\probeagent{} connects diagnosis of observed failures with discovery of further failures
and the patterns they share.

We evaluate \probeagent{} across 22 agent configurations and 10 benchmarks, collecting
1,244 explained failures and producing 536 failure-mode descriptions. These distinguish
specific mistakes within incomplete work, such as planning without acting and stopping
before delivering a file. Human annotators accept 90.0\% of assigned descriptions and
correctly reject 83.3\% of similar alternatives. Review of generated tasks detects
invalid tests that pass execution checks, and enforcing similarity within each group
improves description agreement from 78.0\% to 91.8\% against the model-based reference.
These results demonstrate that \probeagent{} produces usable diagnostic tests and
accurate, specific failure descriptions.

\ifprobeagentpreprint\Needspace{10\baselineskip}\fi
\textbf{Contributions.} Our contributions are:
\begin{enumerate}[leftmargin=*,nosep,topsep=3pt]
\item \textbf{We propose \probeagent{}, an automated framework for discovering and diagnosing
agent failures.} It connects executable testing with failure-mode discovery
(Section~\ref{sec:method}).
\item \textbf{\probeagent{} builds and validates probes to test failure explanations.}
It turns explanations into new executable tests and groups failures by their causes
to generate descriptions supported by examples (Sections~\ref{sec:establish}--\ref{sec:induce}).
\item \textbf{We evaluate \probeagent{} across agents and benchmarks.}
Our experiments demonstrate diverse failure discovery, accurate descriptions, and
effective probe validation
(Section~\ref{sec:experiments}).
\end{enumerate}

\section{Preliminaries}
\label{sec:prelim}

Let $A$ denote an agent. A \emph{rollout} is one execution, producing a trajectory
$z=(o_0,a_1,o_1,\ldots,a_T,o_T)$ of actions $a_t$ and observations $o_t$.

\begin{definition}[Probe]
\label{def:probe}
A probe is an executable test $s=(E,x,g,\pi^\star)$, comprising an environment $E$,
instruction $x$, binary task checker $g$, and reference solution $\pi^\star$.
The checker returns $g(z)=1$ for success and $g(z)=0$ for failure.
\end{definition}

\begin{definition}[Failure]
\label{def:failure}
A failure is a diagnosed error $r=(s,z,u,e)$ from an execution with $g(z)=0$ on a
validated probe $s$. It includes the trajectory $z$, a summary $u$, and a failure explanation
$e$ (Definition~\ref{def:explanation}). One unsuccessful execution may contain several
errors, each represented by its own $r$.
\end{definition}

\begin{definition}[Failure explanation]
\label{def:explanation}
A failure explanation $e=(b,\ell,\tau,\delta)$ identifies a proposed cause of one
failure: the erroneous action or omitted action $b$, the point in the task where it
occurs $\ell$, the task condition $\tau$ under which it arises (the \emph{trigger}),
and a corrective action $\delta$. Each field is supported by the trajectory $z$.
\end{definition}

\begin{definition}[Failure-mode description]
\label{def:description}
A failure-mode description $c$ is a natural-language statement of a shared error.
Write $c(r)\in\{0,1\}$ for the judgment of whether failure $r$ exhibits that error.
A \emph{failure mode} is a tuple $(c,R,B)$ with nonempty supporting failures $R$ and
disjoint contrasting failures $B$, for which the description satisfies
\[
c(r)=1\quad(r\in R),\qquad c(r)=0\quad(r\in B).
\]
A failure explanation $e$ concerns one trajectory; a description $c$ states the pattern used
to group failures. Section~\ref{sec:induce} specifies how these judgments are made.
\end{definition}

\textbf{Example.} An agent computes a total but stops without saving the required
report. The failure $r$ records this error and its trajectory. Its explanation $e$
identifies the omitted save after computation, the requirement to deliver a file,
and the correction of writing that file. A description $c$, ``stops after computing
without delivering the required file,'' can cover this failure and similar errors
in database or spreadsheet tasks. A failed write does not match this description.

\begin{definition}[Sibling environment]
\label{def:sibling}
For a failure $r$ from probe $s$, interpret its trigger $\tau(r)$ as a predicate on
probes: $\tau(r)(s')=1$ means that $s'$ contains the proposed triggering condition.
A sibling is a new probe $s'\ne s$ with $\tau(r)(s')=\tau(r)(s)=1$, constructed by
varying domain, data, tools, naming, or output format. Its environment is a
\emph{sibling environment}. We validate the new probe before testing the same agent $A$ on it
to determine whether the error recurs.
\end{definition}

\textbf{Objective.}
For agent $A$, search policy $\eta$ constructs and tests probes, collecting failures
$\mathcal D_\eta$ at cost $C(\eta)$. Given budget $B_{\mathrm{test}}$ and fixed induction
procedure $\mathcal I_\theta$, we seek the most distinct, supported failure modes:
\begin{equation}
\max_\eta\;\mathbb E\!\left[\left|\mathcal I_\theta(\mathcal D_\eta)\right|\right]
\quad\text{s.t.}\quad C(\eta)\leq B_{\mathrm{test}}.
\label{eq:discovery-goal}
\end{equation}
The expectation accounts for randomness in task construction and agent execution.
The budget covers construction, validation, and testing. We count modes supported by
validated probes and descriptions satisfying Definition~\ref{def:description},
using the same induction settings when comparing search policies.

\section{The \probeagent{} Framework}
\label{sec:method}

\begin{figure}[!t]
  \raggedleft
  \ifprobeagentpreprint
  \includegraphics[width=\linewidth]{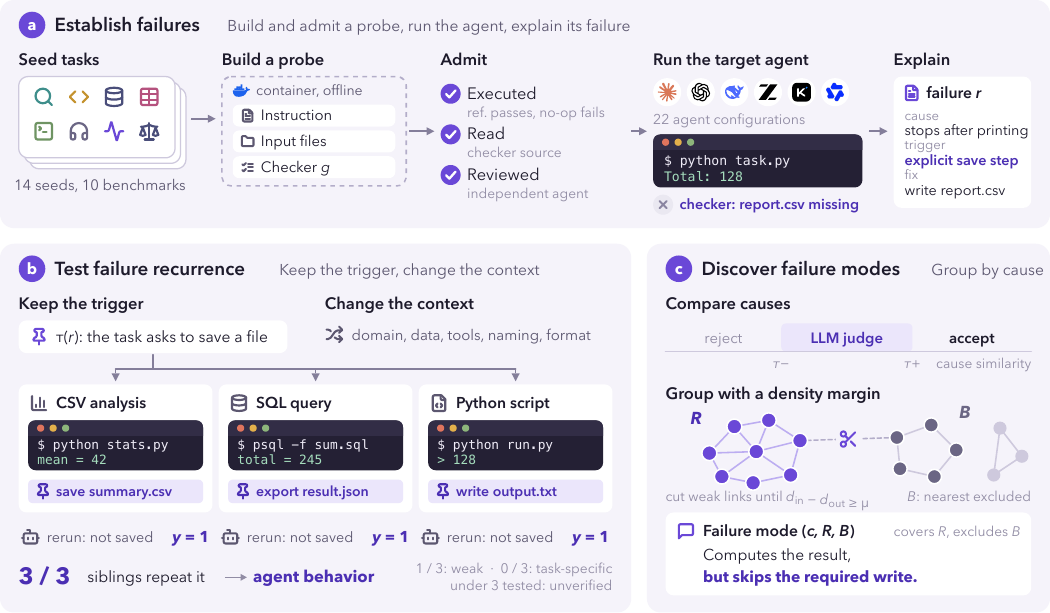}
  \else
  \includegraphics[width=\linewidth]{figures/probeagent_teaser.pdf}
  \fi
  \caption{\textbf{\ifprobeagentpreprint\probeagent{} in detail.\else Overview of \probeagent{}.\fi}
(a)~\textbf{Establish failures:} construct and validate a probe from a benchmark task,
and explain the target's failed trajectory.
(b)~\textbf{Test failure recurrence:} preserve the structural trigger while varying the
context; here, each task computes a result but omits delivery.
(c)~\textbf{Discover failure modes:} use cause-based similarity and density-constrained
grouping to generate descriptions that cover each group and exclude similar errors outside it.
}
\label{fig:probeagent-diagnosis-teaser}
\end{figure}

\probeagent{} connects three stages (Figure~\ref{fig:probeagent-diagnosis-teaser}):
\emph{Establish failures} builds and validates probes, then explains failed trajectories;
\emph{Test failure recurrence} checks whether the explained error appears in new tasks;
and \emph{Discover failure modes} groups failures and describes their shared errors.
Search alternates between the first two stages, and discovery of failure modes uses
the collected failures. Appendix~\ref{app:protocols} specifies the LLM configuration
for construction, review, explanation, and judging.

\subsection{Establish failures}
\label{sec:establish}

To diagnose an agent, we first need tests whose instructions, environment, and
checker agree. We construct these tests from benchmark tasks and validate them
before interpreting the agent's errors.

\textbf{Construct a probe from a seed task.}
A seed task $t$ is a benchmark task used to start the search. We select tasks with
inspectable checkers on which the target agent fails an initial trial. The reference
solution succeeds and inaction fails, as specified by Equation~\eqref{eq:oracle}.
The builder receives the task instructions, input files, and checker, and constructs
$s\sim\operatorname{Build}(t)$. It changes the setting or operational demands while
preserving the question and required outcome. Appendix~\ref{app:protocols} gives
the construction budget and task-preservation checks.

\textbf{Validate the probe.}
We admit a generated probe after three checks.
\textbf{a) Executed: test reference and null controls.} The reference must succeed and
inaction ($\mathrm{nop}$) must fail; $g$ evaluates their executions. Model-judged tasks
record their verdict type.
\textbf{b) Read: inspect checker sources.} We check for inaccessible or external assertions.
\textbf{c) Reviewed: check instruction--implementation consistency.} A separate read-only
agent checks for shortcuts, missing requirements, and solution leakage. It reports evidence
but cannot repair candidates; rejected probes return to the builder within budget.
This review detects shortcuts that execution controls can miss (Section~\ref{sec:validity}).
With binary check results $\lambda_{\mathrm{exec}}$,
$\lambda_{\mathrm{read}}$, and $\lambda_{\mathrm{review}}$, admission is
\begin{equation}
\begin{aligned}
\lambda_{\mathrm{exec}}(s)&=\mathbf 1\{g(\pi^\star)=1\ \land\ g(\mathrm{nop})=0\},\\
\Lambda(s)&=\lambda_{\mathrm{exec}}(s)\lambda_{\mathrm{read}}(s)\lambda_{\mathrm{review}}(s).
\end{aligned}
\label{eq:oracle}
\end{equation}
Only probes with $\Lambda(s)=1$ proceed to target testing.

\textbf{Explain the target trajectory.}
For an admitted probe, let $z\sim\operatorname{Run}(A,s)$ be the target trajectory.
If $g(z)=0$, an LLM analyzes the task, checker feedback, and trajectory to return
summary--explanation pairs
$(u,e)\in\operatorname{Analyze}(s,z)$, yielding
\begin{equation}
\mathcal R(s,z)=\{(s,z,u,e):(u,e)\in\operatorname{Analyze}(s,z)\},\qquad g(z)=0.
\label{eq:analyze}
\end{equation}
Each element is a failure $r$ (Definition~\ref{def:failure}). In
Figure~\ref{fig:probeagent-diagnosis-teaser}(a), the explanation identifies an omitted save,
proposes writing the report, and supplies the trigger for sibling tests.

\subsection{Test failure recurrence}
\label{sec:search}

A failure explanation proposes why one trajectory went wrong. We test that
explanation by constructing tasks in which the same error could occur, while
changing details that are specific to the original task.

\textbf{Construct sibling tasks.}
Given failure $r$ from $s$, the builder samples
$s_j\sim\operatorname{Build}_{\mathrm{sib}}(r,\tau(r))$. We construct and validate each sibling so that
$s_j\ne s$, $\tau(r)(s_j)=1$, and $\Lambda(s_j)=1$ (Definition~\ref{def:sibling}).
For example, CSV analysis can become a JSON database query while retaining the requirement
to save the result (Figure~\ref{fig:probeagent-diagnosis-teaser}(b)).

\textbf{Test the explanation.}
\probeagent{} runs the same agent to obtain $z_{j,t}\sim\operatorname{Run}(A,s_j)$ for attempts
$t=1,\ldots,n_j$. An LLM recurrence judge $H(r,z_{j,t})\in\{0,1\}$ compares the trajectory
with $r$'s failure explanation and corrective action. It returns 1 if the same error
recurs under the specified condition, and 0 otherwise. For $m$ tested sibling
environments, we count the environments in which the error recurs:
\begin{equation}
y_j=\max_{1\leq t\leq n_j}H(r,z_{j,t}),\qquad
\rho=\sum_{j=1}^{m}y_j.
\label{eq:reproduction}
\end{equation}
Thus repeated occurrences in one environment count once. An omitted save matches the
example failure; incorrect computation or an unsuccessful write does not.
With $m_0=3$ and $\theta=2/3$, the test outcome is
\begin{equation}
v(r)=
\begin{cases}
\textsc{agent behavior} & m\geq m_0,\;\rho/m\geq\theta,\\
\textsc{task-specific} & m\geq m_0,\;\rho=0,\\
\textsc{weak} & m\geq m_0,\;0<\rho/m<\theta,\\
\textsc{unverified} & m<m_0.
\end{cases}
\label{eq:verdict-rule}
\end{equation}
A mode pools its members' distinct tests and reproduction counts, counting each test
once and recording conflicting outcomes among its failures.

\textbf{Balance new tasks and sibling tests.}
\probeagent{} alternates between exploring new tasks and testing sibling tasks. During
exploration, \probeagent{} selects the task category with the fewest admitted probes.
Categories are determined by the search goal and seed tasks, and ties are broken
at random.
A newly observed failure label schedules construction of three sibling tasks. After
three consecutive sibling construction attempts, we construct a probe from a seed
task to continue exploring for new failures.
Appendix~\ref{app:protocols} gives the selection rule and budgets.

\subsection{Discover failure modes}
\label{sec:induce}

Different tasks can expose the same error, while similar unsuccessful outcomes
can have different causes. \probeagent{} groups the collected failures by cause,
then generates a description for each group. We call this process \emph{induction}.
It uses failures from both seed-task probes and sibling tasks, including new errors
found during sibling tests, and outputs modes $(c,R,B)$
(Definition~\ref{def:description}; Figure~\ref{fig:probeagent-diagnosis-teaser}(c)).

\textbf{Cause-based similarity.}
For each failure $r=(s,z,u,e)$, we form text $\kappa(r)$ by combining its summary
$u$ with the cause, corrective action, trigger, and task stage from $e$, omitting
absent fields. We embed this text as $\phi(r)=\mathrm{emb}(\kappa(r))$, subtract
the corpus mean $\bar\phi$, and normalize:
\begin{equation}
\psi(r)=\frac{\phi(r)-\bar\phi}{\|\phi(r)-\bar\phi\|_2},
\qquad \mathrm{sim}(r,r')=\langle\psi(r),\psi(r')\rangle .
\label{eq:centre}
\end{equation}
This reduces the shared embedding direction associated with describing agent failures.
We use an LLM as a pairwise judge $J$ that compares two failures' explanations, triggers,
and corrections: $J(r,r')=1$ denotes the same mechanism and $0$ different mechanisms.
It labels a calibration sample spanning the full similarity range. We choose thresholds
$\tau_+,\tau_-$ so that, on this sample, at least $q_+=0.90$ of pairs with
$\mathrm{sim}\geq\tau_+$ receive $J=1$, and at least $q_-=0.98$ of pairs with
$\mathrm{sim}\leq\tau_-$ receive $J=0$. Thus $q_\pm$ specify target agreement rates
for accepting or rejecting a pair as sharing the same error. We accept pairs above
$\tau_+$ and reject pairs below $\tau_-$. We call the interval between these thresholds
the \emph{ambiguous band}. Within this interval, we send pairs among each failure's
$k_{\mathrm{nn}}=15$ nearest neighbors to $J$ and retain those receiving $J=1$.
Other pairs in the interval remain unconnected. Appendix~\ref{app:protocols} gives
the threshold selection rule.

\textbf{Density-constrained grouping.}
Failures form a graph $G=(\mathcal D,E)$ whose edges connect pairs accepted by
the pair-selection procedure above. Each retained edge has weight
$w(r,r')=\mathrm{sim}(r,r')$.
Chains of such links can connect otherwise different failures, so \probeagent{} also checks
each group's overall similarity. For a connected group $K$, let $d_{\mathrm{in}}(K)$
be the mean similarity over all member pairs and $d_{\mathrm{out}}(K)$ the maximum
similarity between any member and any outside failure. For connected partitions
$\mathcal P(G)$ of this graph, our grouping target is
\begin{equation}
\min_{\mathcal K\in\mathcal P(G)}|\mathcal K|
\quad\text{s.t.}\quad d_{\mathrm{in}}(K)-d_{\mathrm{out}}(K)\geq\mu
\quad\forall K\in\mathcal K:\ |K|>1.
\label{eq:cohesion}
\end{equation}
We approximate this objective with a greedy procedure. For a group that violates
the constraint,
we remove edges in ascending weight order until it disconnects, then check each
resulting component again. Splitting continues until each group satisfies the
constraint or contains one failure. The similarity margin $\mu$ controls the
required gap between within-group and outside similarity. Unlike $\tau_\pm$,
which decide individual pair links, $\mu$ checks a whole group. Increasing it
can split a broad group into smaller groups with more specific errors.

\textbf{Generate and check failure-mode descriptions.}
For each group $R$, the \emph{boundary} $B$ contains the nearest $k_{\mathrm{nn}}$
excluded failures, ranked by their maximum similarity to a failure in $R$. These are failures
that look similar but belong to other groups. An LLM generates a description $c$, also
called a \emph{claim}. We use the induction LLM as a judge to check whether the error and condition
stated in $c$ agree with each failure's explanation and corrective action. We accept
$c$ when it covers every failure in $R$ and excludes every failure in $B$ (Definition~\ref{def:description}); otherwise the group remains unresolved.
For a new failure $r$, the corresponding matching modes are $\mathcal M(r)=\{k:c_k(r)=1\}$.
Thus a failure belongs to a mode when it exhibits the error stated in its description.
A failure may match multiple
descriptions or remain unassigned. Discovery groups partition the corpus; an execution
with several diagnosed errors can contribute to several modes.

We evaluate these descriptions on failures withheld before induction
(Section~\ref{sec:granularity}). Algorithm~\ref{alg:induce} gives the full procedure;
Appendix~\ref{app:protocols} specifies the model roles and scoring rules.

\section{Experiments}
\label{sec:experiments}

\subsection{Experimental setup}

\textbf{Models and harnesses.}
We evaluate 22 configurations combining 15 models and seven harnesses.
The models include GPT-5.6 and GPT-5.4-nano \citep{openai2026gpt56,openai2026gpt54nano},
Claude \citep{anthropic2026claude}, DeepSeek-V4-Pro \citep{deepseek2026v4},
Kimi K3 \citep{moonshot2026kimik3}, GLM-5 and GLM-5.2 \citep{zai2026glm5,zai2026glm52},
Qwen3.8-Max \citep{qwen2026qwen38max}, Grok~4.5 \citep{xai2026grok45},
and MiniMax-M2.7 \citep{minimax2026m27}.
The harnesses are Terminus~2 \citep{harbor2026terminus}, Codex \citep{openai2026codex},
Claude Code \citep{anthropic2026claudecode}, Goose \citep{goose2026},
opencode \citep{anomaly2026opencode}, OpenHands \citep{wang2025openhands},
and Aider \citep{aider2026}. Table~\ref{tab:profile} lists the exact model--harness pairings.
Appendix~\ref{app:protocols} specifies the internal model configuration for
construction, review, analysis, and induction.

\textbf{Tasks and protocol.}
We start from 14 seed tasks from 10 benchmarks spanning retrieval, code editing,
performance engineering, data pipelines, terminal operation, tabular and spreadsheet work,
clinical tasks, customer service, and law (Appendix~\ref{app:corpus}).
Each diagnostic search constructs and validates new probe environments from these
tasks, then collects failures from the agent's trajectories. We group these failures
and generate descriptions of their shared errors. For corpus analysis, we use the group-level
similarity margin $\mu=0.02$ defined in Equation~\eqref{eq:cohesion}; ablations
examine other values.
Probes run as Harbor tasks \citep{harbor2026} in Docker containers \citep{docker2026}.
Search uses three rollouts per environment.
Appendix~\ref{app:hyper} gives the hyperparameters and execution details.

\subsection{Failure discovery across agents and tasks}
\label{sec:profile}
\label{sec:taxonomy}

\textbf{\probeagent{} discovers diverse failure modes across agents and tasks.}
We run \probeagent{} across 347 generated environments and collect 1,244 explained failures
from which we generate 536 failure-mode descriptions. Table~\ref{tab:modes} provides examples of the
resulting descriptions, including treating a human transfer as completion,
planning an action without executing it, and continuing research without delivering
a file. Each description specifies an error in its task context.

\begin{table}[!htbp]
\caption{Examples of induced claims, abridged for presentation. $|R|$ is the number of supporting failures.}
\label{tab:modes}
\small
\setlength{\tabcolsep}{3pt}
\zebrabody
\ifprobeagentpreprint
\def\descriptionwidth{1.40}\def\modelswidth{0.60}
\setlength{\tabcolsep}{5pt}
\else
\def\descriptionwidth{1.62}\def\modelswidth{0.38}
\fi
\begin{tabularx}{\linewidth}{@{}>{\hsize=\descriptionwidth\hsize}Lr>{\hsize=\modelswidth\hsize}L@{}}
\preprinttabletop
\headrow
\thead{Failure description} & \thead{$|R|$} & \multicolumn{1}{H}{\thead{Models}} \\
\preprinttablemid
The checker rejects the submission when execution leaves a generated artifact or modified file
  outside the permitted delivery surface.
  & 11 & \modelname{cl}{Sonnet 4.5}, \modelname{oa}{GPT-5.4-nano} \\
After a successful human transfer, the agent treats the transfer as sufficient termination and
  omits the required closing actions.
  &  8 & \modelname{oa}{GPT-5.6-sol}, \modelname{ds}{DeepSeek-V4}, \modelname{ms}{Kimi K3} \\
Reaching a usable startup point, the agent remains at \emph{planning} the required call instead
  of issuing it, and the run ends with nothing created.
  &  5 & \modelname{oa}{GPT-5.4-nano}, \modelname{za}{GLM-5}, \modelname{za}{GLM-5.2} \\
When the deliverable is a JSON file, the agent stays in research until the run ends and never
  creates it.
  &  4 & \modelname{xa}{Grok 4.5}, \modelname{qw}{Qwen3.8-Max} \\
\preprinttablebottom
\end{tabularx}
\end{table}

\ifprobeagentpreprint
Table~\ref{tab:profile} summarizes these discoveries across all 22 configurations.
\begin{table}[!tb]
\caption{\textbf{Failure discoveries by agent configuration.} Shared groups contain failures
from multiple configurations; largest gives the size of the largest group containing
a failure from that row. Group counts can overlap across rows.}
\label{tab:profile}
\fontsize{9.3}{11.5}\selectfont
\setlength{\tabcolsep}{4pt}
\renewcommand{\arraystretch}{1.10}
\setlength{\extrarowheight}{1.0pt}
\newcommand{\profilecount}[1]{#1}
\begin{tabularx}{\linewidth}{@{}>{\hsize=1.12\hsize}L>{\hsize=0.88\hsize}Lrrrr@{}}
\preprinttabletop
\headrow
\thead{Model} & \thead{Harness} & \thead{Failures} & \thead{Groups} & \thead{Shared} & \thead{Largest} \\
\preprinttablemid
\rowcolor{tblZebra}
\modelname{ds}{DeepSeek-V4-Pro} & \harnessname{t2}{\textcolor{ckMuted}{Terminus 2}} & \profilecount{132} & 68 & 7 & 8 \\
\modelname{oa}{GPT-5.6-sol} & \harnessname{t2}{\textcolor{ckMuted}{Terminus 2}} & \profilecount{132} & 58 & 4 & 8 \\
\rowcolor{tblZebra}
\modelname{cl}{Claude Sonnet 5} & \harnessname{t2}{\textcolor{ckMuted}{Terminus 2}} & \profilecount{105} & 56 & 5 & 4 \\
\modelname{cl}{Claude Sonnet 4.5} & \harnessname{t2}{\textcolor{ckMuted}{Terminus 2}} & \profilecount{143} & 52 & 7 & 15 \\
\rowcolor{tblZebra}
\modelname{ms}{Kimi K3} & \harnessname{t2}{\textcolor{ckMuted}{Terminus 2}} & \profilecount{87} & 52 & 6 & 8 \\
\modelname{xa}{Grok 4.5} & \harnessname{t2}{\textcolor{ckMuted}{Terminus 2}} & \profilecount{84} & 47 & 3 & 5 \\
\rowcolor{tblZebra}
\modelname{za}{GLM-5} & \harnessname{t2}{\textcolor{ckMuted}{Terminus 2}} & \profilecount{72} & 35 & 0 & 4 \\
\modelname{oa}{GPT-5.6-luna} & \harnessname{t2}{\textcolor{ckMuted}{Terminus 2}} & \profilecount{64} & 33 & 4 & 8 \\
\rowcolor{tblZebra}
\modelname{oa}{GPT-5.4-nano} & \harnessname{t2}{\textcolor{ckMuted}{Terminus 2}} & \profilecount{73} & 30 & 3 & 12 \\
\modelname{cl}{Claude Opus 5} & \harnessname{t2}{\textcolor{ckMuted}{Terminus 2}} & \profilecount{55} & 24 & 1 & 4 \\
\rowcolor{tblZebra}
\modelname{oa}{GPT-5.6-terra} & \harnessname{t2}{\textcolor{ckMuted}{Terminus 2}} & \profilecount{50} & 21 & 2 & 6 \\
\modelname{qw}{Qwen3.8-Max} & \harnessname{t2}{\textcolor{ckMuted}{Terminus 2}} & \profilecount{39} & 20 & 5 & 5 \\
\rowcolor{tblZebra}
\modelname{mm}{MiniMax-M2.7} & \harnessname{t2}{\textcolor{ckMuted}{Terminus 2}} & \profilecount{35} & 15 & 2 & 15 \\
\modelname{cl}{Claude Opus 4.8} & \harnessname{t2}{\textcolor{ckMuted}{Terminus 2}} & \profilecount{11} & 8 & 2 & 15 \\
\rowcolor{tblZebra}
\modelname{za}{GLM-5.2} & \harnessname{t2}{\textcolor{ckMuted}{Terminus 2}} & \profilecount{11} & 6 & 1 & 4 \\
\addlinespace[3pt]\midrule[0.3pt]\addlinespace[2pt]
\modelname{cl}{Claude Sonnet 5} & \harnessname{oh}{\textcolor{ckMuted}{OpenHands}} & \profilecount{40} & 20 & 2 & 5 \\
\rowcolor{tblZebra}
\modelname{cl}{Claude Sonnet 5} & \harnessname{go}{\textcolor{ckMuted}{Goose}} & \profilecount{35} & 16 & 3 & 6 \\
\modelname{cl}{Claude Sonnet 4.5} & \harnessname{cc}{\textcolor{ckMuted}{Claude Code}} & \profilecount{37} & 16 & 0 & 3 \\
\rowcolor{tblZebra}
\modelname{cl}{Claude Sonnet 5} & \harnessname{oc}{\textcolor{ckMuted}{opencode}} & \profilecount{30} & 14 & 2 & 6 \\
\modelname{oa}{GPT-5.6-sol} & \harnessname{cx}{\textcolor{ckMuted}{Codex}} & \profilecount{9} & 7 & 1 & 4 \\
\rowcolor{tblZebra}
\modelname{oa}{GPT-5.4-nano} & \harnessname{cx}{\textcolor{ckMuted}{Codex}} & \profilecount{0} & 0 & 0 & 0 \\
\modelname{cl}{Claude Sonnet 5} & \harnessname{ad}{\textcolor{ckMuted}{Aider}} & \profilecount{0} & 0 & 0 & 0 \\
\midrule[0.3pt]
\multicolumn{2}{@{}l}{\textbf{Corpus}} & \textbf{1{,}244} & \textbf{565} & \textbf{27} & \textbf{15} \\
\preprinttablebottom
\end{tabularx}
\end{table}

\FloatBarrier
\fi

\begin{figure}[!tb]
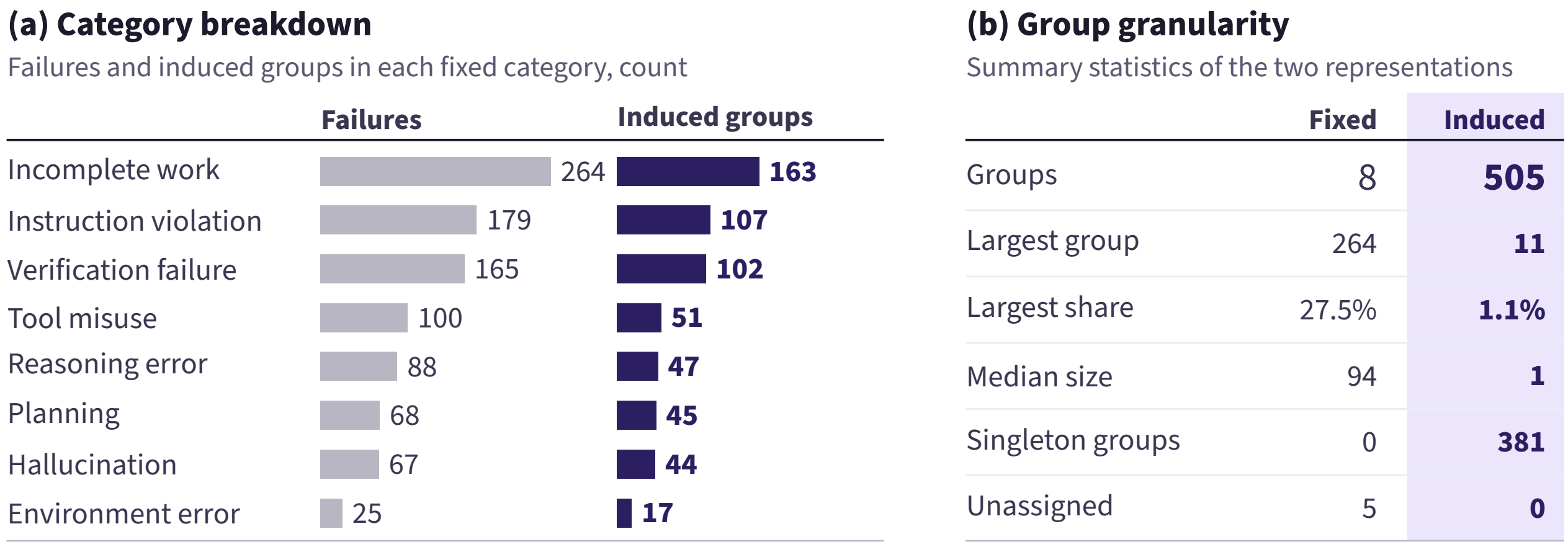

\centering
\FigTaxonomy
\caption{\textbf{Fixed categories and induced descriptions} on 961 failures.
Left: failures and induced groups represented in each category.
Right: summary statistics for the two representations. The taxonomy provides broad
coverage, while induction separates more specific mechanisms.}
\label{fig:taxonomy}
\end{figure}

\textbf{\probeagent{} distinguishes failures within broad error categories.}
We compare our groups with a baseline that assigns failures to eight fixed categories,
including incomplete work, instruction violations, and verification failures
(Figure~\ref{fig:taxonomy}). Both use the same 961 failures. The \emph{incomplete} category contains 264 failures
(27.5\% of the set), spanning 163 induced groups. The pattern extends to
\emph{instruction violation} and \emph{verification failure}, which contain
107 and 102 induced groups, respectively. Thus the distinctions found by \probeagent{}
occur across several common error categories, rather than only within one type
of unsuccessful outcome.

The largest induced group contains 11 failures, compared with 264 in the largest
fixed category (Figure~\ref{fig:taxonomy}, right). A category such as \emph{incomplete}
identifies an outcome; our descriptions distinguish the actions or omissions behind it.

\ifprobeagentpreprint
\Needspace{8\baselineskip}
\textbf{Related agent configurations share more discovered failure modes.}
We keep four seed tasks and the search settings fixed, and form nine related pairs:
three model-version pairs under Terminus~2 and six harness pairs using Sonnet~5.
Four of these nine pairs (44.4\%) share at least one induced group, compared with
26 of the other 181 pairs (14.4\%; Figure~\ref{fig:pairshare}). These shared groups
identify specific errors that persist across a change in model or harness.
Appendix~\ref{app:comparison} gives the configurations and individual overlaps.

\begin{figure}[!htb]
\centering
\FigPairShare
\caption{\textbf{Failure sharing across configuration pairs.}
\ifprobeagentpreprint
Bars show the fraction of pairs sharing at least one induced group: $4/9$ related
pairs and $26/181$ other pairs. Related pairs differ only in model version or harness.
\else
Each square represents a pair and is filled when the configurations share at least
one group. The designed pairs share at a rate of $4/9$; other pairs share at $26/181$.
\fi}
\label{fig:pairshare}
\end{figure}

\textbf{Group counts differ within these pairs.}
Moving from GLM-5 to GLM-5.2 under Terminus~2 lowers the number of distinct groups
from 35 to 6, while moving Sonnet~4.5 from Claude Code to Terminus~2 raises it from
16 to 52 (Figure~\ref{fig:pairs}). These counts describe adaptive discovery campaigns,
not matched comparisons; Section~\ref{sec:recurrence} reports matched replay.

\begin{figure}[!tb]
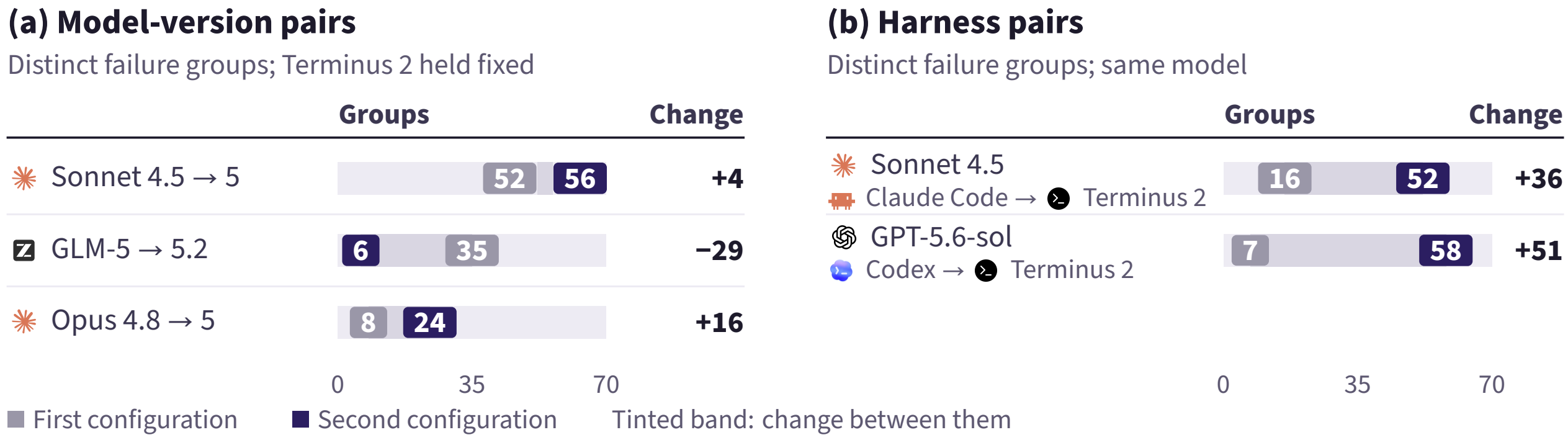

\centering
\FigPairs
\caption{\textbf{Group counts for selected configurations.}
(a) Model-version pairs under Terminus~2.
(b) Harness pairs with the same model. Counts match Table~\ref{tab:profile}; Appendix~\ref{app:comparison} gives the setup.
Grey and navy labels give the first and second configurations; the tinted band spans the
change.}
\label{fig:pairs}
\end{figure}
\FloatBarrier
\fi

\subsection{Accuracy of failure-mode induction}
\label{sec:granularity}

\textbf{\probeagent{} accurately characterizes failure modes.}
We evaluate description accuracy by testing whether each description correctly identifies
matching failures. The induction judge assigns held-out failures using the descriptions, and
the pairwise judge confirms each assignment against the selected group's first
member in corpus order. The classifier may return no match; agreement is measured
among assigned failures. With 25\% of failures held
out before induction, \probeagent{} achieves 88.5\% agreement on the 1,244-failure corpus at
$\mu=0.02$. These failures are excluded from centering, calibration, and description
construction. The result supports the accuracy of the induced failure descriptions
(Appendix~\ref{app:induction}).

\textbf{Human evaluation supports description accuracy and specificity.}
We ask two annotators to evaluate 30 failures, each paired with its assigned group's
description and a similar alternative from another group. Each annotator evaluates
all 60 pairs. They accept 28 and 26 assigned descriptions, respectively, and each
correctly rejects 25 alternatives, giving mean acceptance and rejection rates of
90.0\% and 83.3\%. These judgments support accurate descriptions that distinguish
similar errors (Appendix~\ref{app:human-eval}).

\subsection{Effectiveness of probe validation}
\label{sec:validity}

\textbf{Review of task requirements detects defects that execution checks miss.}
Across 1,451 build rounds, the execution gate rejects 349 candidates, and the reviewer
rejects a further 638 after executable controls pass. The most common finding is a
rewarded shortcut that bypasses the intended task, observed in 565 rounds. Neither the reference nor null run attempts these shortcuts. In 323 rounds, the checker does not evaluate the requested
outcome; in 255, it imposes an unstated requirement. Findings can overlap within a round. These findings show why both checks contribute to probe validation: execution
tests whether the reference succeeds and inaction fails, while semantic review
examines whether the checker rewards the work specified in the instruction.
\ifprobeagentpreprint
Table~\ref{tab:review-findings} gives the complete finding counts. Review also detects
missing instructions and exposed solutions, which change the task the agent is
asked to solve. Appendix~\ref{app:gates} reports the sequential gate counts.

\begin{table}[!htbp]
\caption{\ifprobeagentpreprint
Semantic-review findings over 1,451 build rounds. A round may receive multiple findings.
\else
Semantic-review findings. Categories may overlap within a build round.
\fi}
\label{tab:review-findings}
\small\zebrabody
\begin{tabularx}{\linewidth}{@{}Lr@{}}
\preprinttabletop
\headrow
\thead{Finding} & \thead{Rounds} \\
\preprinttablemid
A shortcut receives reward without completing the intended work & 565 \\
The checker does not evaluate the requested outcome & 323 \\
The checker imposes an unstated requirement & 255 \\
The instruction omits a rule needed to solve the task & 220 \\
The instruction cannot be followed as written & 92 \\
A task claim lacks an implementing mechanism & 81 \\
Checker assertions are inaccessible to the reviewer & 59 \\
The environment leaks its solution & 51 \\
\preprinttablebottom
\end{tabularx}
\end{table}

\else
Appendix~\ref{app:gates} gives the complete gate and finding counts.
\fi

Classification of 940 failures by whether they concern task completion, the harness,
or formatting identifies 95.0\% as substantive
failures, 3.1\% as harness-related problems, and 1.9\% as format-related problems.
Reference solutions pass in all nine audited environments where every tested
agent failed. Appendix~\ref{app:validity} reports these checks together with the
manual environment audit, including expected-answer errors and format ambiguities.

\subsection{Sibling testing separates recurring and task-specific failures}
\label{sec:recurrence}

\textbf{Sibling tests distinguish recurring behavior from context-specific failures.}
Figure~\ref{fig:funnel}\ifprobeagentpreprint(a)\fi{} summarizes 239 proposed failure explanations, of which 119 reached
sibling testing within the search budget. Among those tested, 48 (40.3\%) reproduced in no
sibling, 20 (16.8\%) met the agent-behavior criterion, and 26 (21.8\%) reproduced weakly.
The remaining 25 (21.0\%) had fewer than three tested siblings. The tests give each explanation an
explicit recurrence outcome under changed task contexts.

\begin{figure}[!htb]
\centering
\ifprobeagentpreprint
\FigRecurrenceCost
\caption{\textbf{Failure recurrence and replay cost.}
(a) Sibling outcomes for 119 of 239 explanations; recurrence is counted per
environment (Equation~\eqref{eq:verdict-rule}).
(b) Mean replay input tokens per rollout versus failure rates on 18 matched trials per configuration
(log token axis). IDs refer to the key; circles denote Terminus~2 and diamonds
denote Codex or Claude Code.}
\label{fig:cost}
\else
\FigFunnel
\caption{\textbf{Failure recurrence across sibling tasks.} Of 239 proposed explanations, 119 reached sibling
testing. Verdicts follow Equation~\eqref{eq:verdict-rule}; reproduction is counted per
environment. }
\fi
\label{fig:funnel}
\end{figure}

The tests distinguish explanations that recur across new tasks from those that do
not recur in the tested contexts. This gives each explanation evidence beyond its
original trajectory and identifies recurring errors for further investigation.
Appendix~\ref{app:evidence} links these outcomes to the discovered failure modes.

\ifprobeagentpreprint
\textbf{Replay cost varies across configurations.}
Mean input tokens per rollout span 112k--3.49M across 15 configurations
(Figure~\ref{fig:cost}(b)); Appendix~\ref{app:comparison} details the replay protocol.

\textbf{Matched replay compares failure rates on fixed tasks.}
With 18 matched trials per configuration on five environments, failure rates range
from 22.2\% for Kimi~K3 to 94.4\% for GPT-5.4-nano, both under Terminus~2
(Figure~\ref{fig:replay}). The intervals are wide at this sample size;
Appendix~\ref{app:comparison} gives the paired tests.

\begin{figure}[!tb]
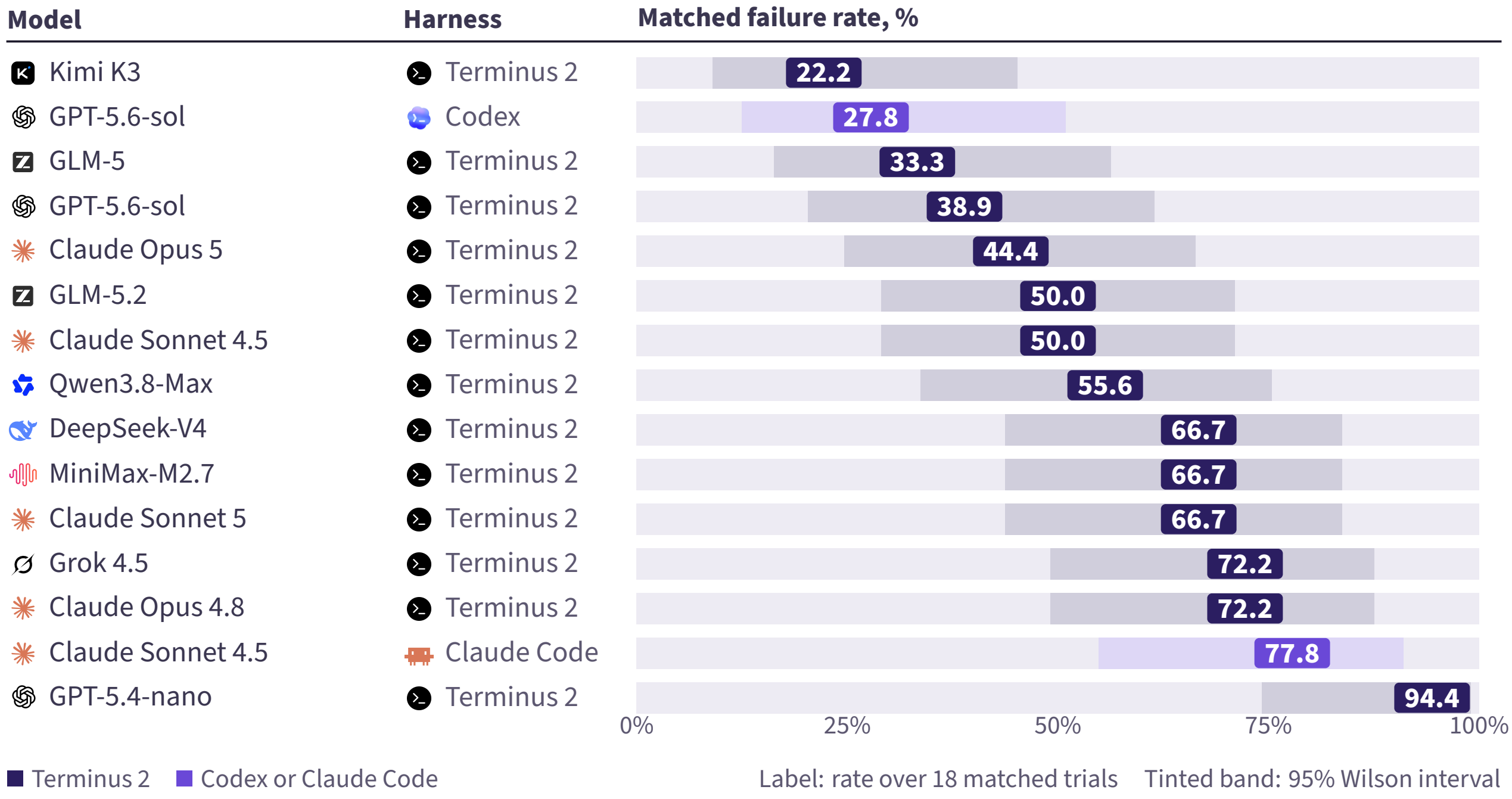

\centering
\FigReplay
\caption{\textbf{Matched replay results.} Rates use 18 matched trials per valid
configuration across five environments; intervals are 95\% Wilson intervals,
shown as tinted bands with each rate printed at its estimate.
The pool was selected for discriminating failures. Infrastructure-only executions
are excluded.}
\label{fig:replay}
\end{figure}
\fi

\ifprobeagentpreprint\FloatBarrier\Needspace{9\baselineskip}\fi
\subsection{Grouping components and efficiency}
\label{sec:ablation}

We test which grouping choices improve description accuracy and reduce LLM calls.

\textbf{The group-level constraint prevents excessive merging.}
\ifprobeagentpreprint Figure~\ref{fig:ablation}\else Table~\ref{tab:ablation}\fi{} compares induction variants on the same failure set. Removing the
constraint reduces 505 groups to 23, including a component containing 650 of 961 failures,
and lowers held-out agreement from 91.8\% to 78.0\%. Pairwise links alone therefore
merge failures that the group-level constraint separates.

\ifprobeagentpreprint
\begin{figure}[!htb]
\centering
\FigAblation
\caption{\textbf{Induction ablations} on 961 failures with similarity margin $\mu=0.05$.
Held-out agreement is confirmed by $J$; $\Delta$ gives its change from the full method
in percentage points. Judge calls count pairs sent to $J$. Shared counts groups
spanning agent configurations.}
\label{fig:ablation}
\end{figure}

\else
\begin{table}[!htb]
\caption{\textbf{Induction ablations} on 961 failures with similarity margin $\mu=0.05$.
Shared counts groups spanning agent configurations; judged counts pairs sent to $J$.
Agreement measures held-out assignments confirmed by $J$.}
\label{tab:ablation}
\centering\small
\setlength{\tabcolsep}{5pt}
\zebrabody
\begin{tabularx}{\linewidth}{@{}Lrrrr@{}}
\preprinttabletop
\headrow
\thead{Variant} & \thead{Groups} & \thead{Shared} & \thead{Judged} & \thead{Agree.} \\
\preprinttablemid
Full method & 505 & 11 & 7,449 & 91.8\% \\
No density constraint & 23 & 4 & 7,449 & 78.0\% \\
Fixed thresholds & 500 & 11 & 6,789 & 90.8\% \\
Neighbors $k_{\mathrm{nn}}=5$ & 506 & 10 & 2,614 & 93.3\% \\
Neighbors $k_{\mathrm{nn}}=30$ & 505 & 11 & 14,297 & 92.4\% \\
Single threshold & 500 & 11 & 0 & 92.1\% \\
\preprinttablebottom
\end{tabularx}
\end{table}
\fi

\ifprobeagentpreprint
\textbf{Repeated evaluations retain the accuracy gain from the density constraint.}
On the 961-failure corpus, two evaluations yield held-out agreement of 91.8\% and
91.4\% with the constraint, compared with 78.0\% and 70.7\% without it. The
940-failure audit set yields 91.1\% agreement. The constraint improves agreement
in both evaluations, supporting its role in separating distinct errors.

\fi
\textbf{The similarity margin controls how broadly failures are grouped.}
Reducing $\mu$ from 0.05 to 0.02 increases shared groups from 13 to 27 on the
1,244-failure corpus, while held-out agreement changes from 89.3\% to 88.5\%.
We use $\mu=0.02$ for the main corpus analysis because it identifies more groups
shared across agent configurations with similar agreement. The component ablations
in \ifprobeagentpreprint Figure~\ref{fig:ablation}\else Table~\ref{tab:ablation}\fi{} use $\mu=0.05$; \ifprobeagentpreprint
The sweep in Figure~\ref{fig:granularity}(a,b) shows how increasing the margin separates
failures into more groups while reducing support from multiple failures.
\else
Appendix~\ref{app:induction} reports the full margin sweep.
\fi
\ifprobeagentpreprint
At $\mu=0.20$, only 38 of the 665 groups contain multiple failures.
\begin{figure}[!htb]
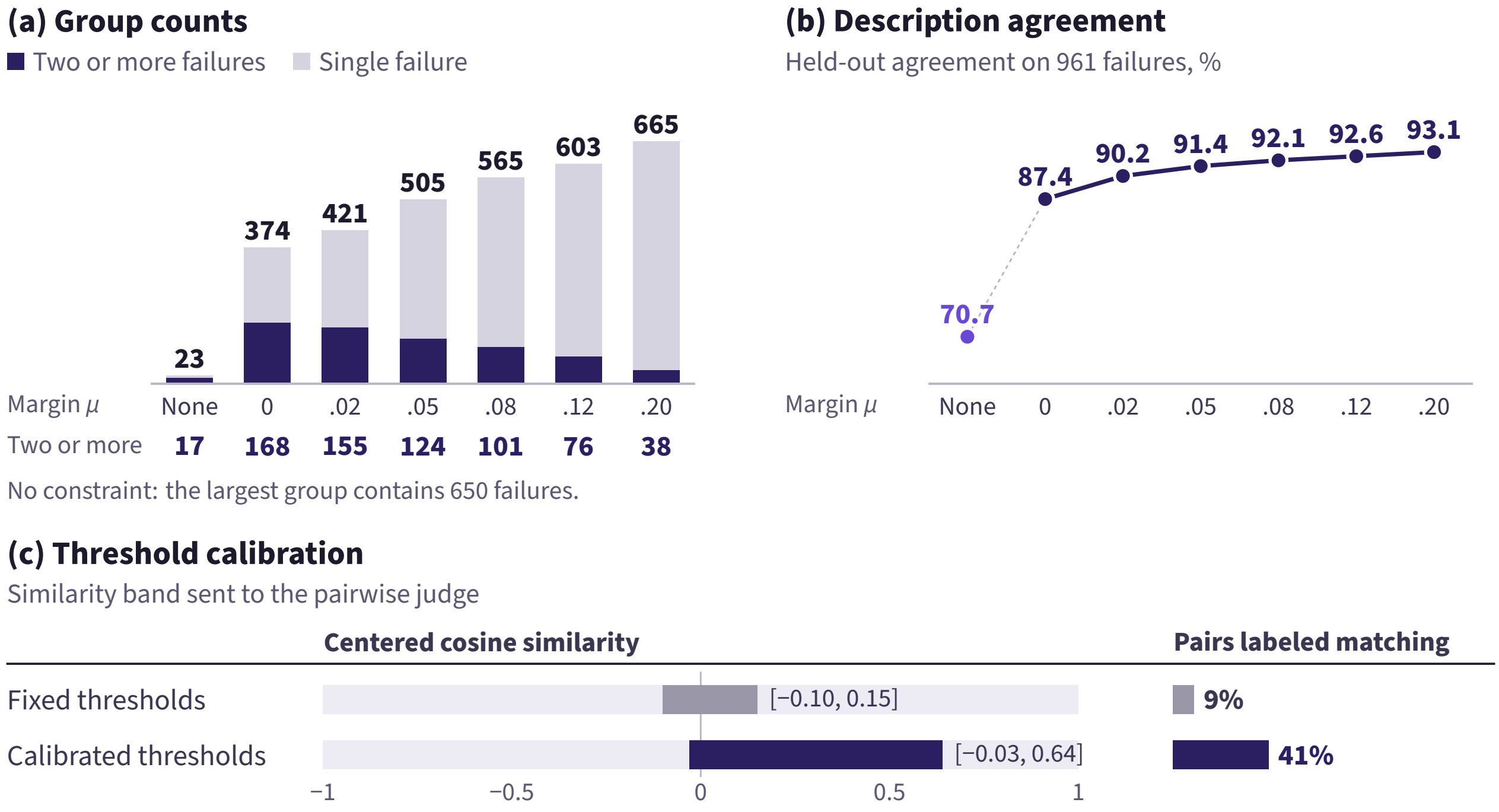

\centering
\FigGranularity
\ifprobeagentpreprint
\par\vspace{5pt}
\FigCalibration
\fi
\caption{\ifprobeagentpreprint
\textbf{Grouping sensitivity and threshold calibration.}
(a) Groups with two or more failures and single-failure groups, with totals above the bars;
(b) held-out description agreement on 961 failures.
(c) Similarity ranges sent to the pairwise judge and the fraction labeled as matching.
Larger margins split groups; calibration concentrates judging on likely matches.
\else
\textbf{Effect of the density margin} on 961 failures. Top: all groups and groups
with at least two failures. Bottom: held-out membership agreement. Larger margins
progressively split groups and increase the singleton share.
\fi}
\label{fig:granularity}
\ifprobeagentpreprint\label{fig:calib}\fi
\end{figure}

\fi

\ifprobeagentpreprint\Needspace{8\baselineskip}\fi
\textbf{Threshold selection and pair reuse reduce judging costs.}
\ifprobeagentpreprint
Selecting thresholds from judge-labeled pairs increases the fraction of matching
pairs in the ambiguous band from 9\% to 41\% (Figure~\ref{fig:calib}(c)).
\else
Selecting thresholds from judge-labeled pairs increases the fraction of pairs judged
to match within the ambiguous band from 9\% to 41\%, compared with fixed thresholds.
\fi
Nearest-neighbor selection reduces judged pairs by 75.5\% while preserving all 53
groups in a 182-failure experiment. Reusing cached pair judgments answers 7,150 of
8,655 ambiguous-band pairs in the 1,244-failure run. Appendix~\ref{app:induction}
reports the similarity analysis, thresholds, and additional cost measurements.

\FloatBarrier
\section{Related Work}

\textbf{Agent evaluation.}
Agent benchmarks cover repository repair, web interaction, desktop operation,
and domain-specific tasks
\citep[e.g.,][]{jimenez2024swebench,zhou2024webarena,xie2024osworld,yao2024taubench,jiang2025medagentbench,liu2024agentbench}.
Recent work also examines how agents progress through a task. For example,
AgentBoard measures intermediate progress, $\tau^2$-Bench studies coordination
when users and agents share control, and Gaia2 evaluates dynamic, asynchronous
environments \citep{ma2024agentboard,barres2025tau2bench,froger2026gaia2}.
Agent-as-a-Judge assesses intermediate work against task requirements, while
AgentRewardBench compares automatic trajectory evaluations with expert
assessments \citep{zhuge2025agentasajudge,lu2025agentrewardbench}.
\ifprobeagentpreprint
Newer benchmarks test agents on longer and more specialized work.
AutoLab evaluates long-horizon research and engineering tasks in which agents
improve a working baseline through repeated experiments \citep{xu2026autolab}.
RSI Arena studies agents that post-train a base language model and then improve it
with human preference votes \citep{huang2026rsiarena}. JobBench grades workflows that
experts in 35 occupations want to delegate \citep{li2026jobbench}.
In specialized domains, NARRA-Gym logs full interactive story episodes to evaluate
narrative agents \citep{huang2026narragym}, and ChemCost scores chemical cost
reasoning against exact ground truth, with stage-level diagnosis of grounding,
retrieval, procurement, and arithmetic failures \citep{wu2026pricereaction}.
\fi
\probeagent{} uses failed tasks to construct further tests, checking whether the
same error recurs under changed task conditions.

\textbf{Task generation and benchmark validity.}
Generated environments expand the experience available for agent training.
For example, AgentGen synthesizes environments and planning tasks, while SWE-smith and
R2E-Gym construct executable software-engineering tasks through repository
modifications, test generation, and commit-based synthesis
\citep{hu2025agentgen,yang2025swesmith,jain2025r2egym}.
\ifprobeagentpreprint
LLM-driven data synthesis also produces training and evaluation data in other settings.
DataGen combines attribute-guided generation with code-based label checks and
retrieval-based fact checks \citep{huang2025datagen}. ChemOrch synthesizes chemistry
instructions with tool-grounded responses, and uses them both to train models and to
build tasks that reveal model weaknesses \citep{huang2025chemorch}. PersonaGym simulates
multi-turn user interactions with evolving preferences for personalization
\citep{ma2026personagym}, and AuraGen injects labeled risks into synthesized agent
trajectories to train the Safiron guardrail \citep{huang2026safiron}.
\fi
Reliable verification is central to using such tasks. A different stream focuses on human validation and
benchmark audits to identify ambiguous specifications, weak tests, and exploitable
evaluation logic
\citep{openai2024swebenchverified,aleithan2024swebenchplus,wang2026benchjack}.
\probeagent{} uses task generation for diagnosis: it constructs validated tests
around a proposed cause of failure and checks whether the error recurs.

\textbf{Automated probing and failure analysis.}
Dynamic evaluation moves beyond fixed test sets by adapting what is tested.
For instance, Meta Probing Agents transform evaluation problems along cognitive dimensions
\citep{zhu2024metaprobing}; AutoDetect coordinates examiners, questioners, and assessors
to expose model weaknesses \citep{cheng2024autodetect};
AutoBencher optimizes dataset descriptions for criteria such as difficulty and novelty
\citep{li2025autobencher};  ACE selects informative capabilities through active
learning \citep{afkanpour2025ace}; and FACT-AUDIT adapts fact-checking assessments to model
responses and evaluates both verdicts and justifications \citep{lin2025factaudit}.
For agents, MAST organizes multi-agent failures into a taxonomy, and Who\&When
provides annotations of the responsible agent and decisive step
\citep{cemri2025mast,zhang2025whowhen}.
\ifprobeagentpreprint
Failures can also emerge from the interaction among agents. In generative multi-agent
systems, collusion-like coordination and conformity arise often across repeated trials
of resource competition, sequential handoff, and collective decisions
\citep{huang2026emergent}. In enterprise workflows, agents that each act within their
own permissions can jointly violate organization-wide constraints; EAIR tests these
risks in executable micro-worlds and uses execution traces to locate and repair them
\citep{ma2026eair}.
\fi
AgentDebug combines critical-error detection
with feedback for later rollouts, while AgentRx uses stepwise constraints to ground
failure localization \citep{zhu2025agentdebug,barke2026agentrx}.
These methods organize or locate errors in observed trajectories. ProbeLLM
\citep{huang2026probellm} discovers failure modes of LLM responses through active
probing and verifiable test cases. \probeagent{} extends this approach to agents:
it constructs and validates environments, uses trajectory explanations to design
new tasks, tests whether the same error recurs, and describes shared failure modes.

\section{Conclusion}

\probeagent{} provides an automated framework for discovering and diagnosing agent failure
modes. Across 22 agent configurations and 10 benchmarks, it collects 1,244 failures
and produces 536 failure-mode descriptions. Its experiments demonstrate diverse
failure discovery, accurate descriptions, and effective probe validation and grouping.
By combining executable tests with tests of recurrence, \probeagent{} helps identify specific
agent errors and the task conditions in which they arise.

\label{sec:main-end}
\clearpage
\subsection*{AI use statement}
Generative models are components of the proposed methodology: they construct probe and
sibling environments, review candidate tasks, explain failed trajectories, and judge failure
pairs and failure claims, as described in Section~\ref{sec:method}. Generative AI tools also
assisted with manuscript restructuring and language revision, consistency checks, LaTeX
formatting, and the design and revision of the overview figure and its description.
\bibliographystyle{plainnat}
\bibliography{refs}
\clearpage
\appendix
\section{Implementation and Detailed Protocols}
\label{app:protocols}

\textbf{Internal model configuration.}
The provided configuration assigns GPT-5.6-sol to probe construction, semantic
review, trajectory analysis, pairwise judging, and claim generation. The reviewer
uses the builder's model settings in a separate read-only session. The induction
judge also classifies held-out failures and checks their assigned groups.
The embedding configuration uses \texttt{google/gemini-embedding-2}.
Builder and reviewer calls use the identifier \texttt{openai/gpt-5.6-sol};
analysis and induction use \texttt{gpt-5.6-sol}. Generative roles default to
temperature zero. These internal roles are configured separately from the target
agents listed in Table~\ref{tab:profile}.

\subsection{Probe construction and search}

The implementation supports discovery, corpus consolidation, and controlled
comparison. The builder runs on the host; target probes use Harbor task directories
and run in Docker. Settings are specified
through configuration files and per-run overrides.

For a benchmark task $t$, seed selection requires
\begin{equation}
\Gamma(t)=\gamma_{\mathrm{read}}(t)\wedge\gamma_{\mathrm{bite}}(t)
\wedge\gamma_{\mathrm{sound}}(t).
\label{eq:seedgate}
\end{equation}
The gates cost zero, one, and two rollouts in order; rejection stops further evaluation.
The reviewability gate checks the available verification sources, including whether
assertions are external or hidden behind an apparent exact-match check. The failure gate
runs the target once and requires $g_t(A(t))<1$. The soundness gate requires a passing
reference and a failing null run. Tasks whose evaluation requires a model judge are
identified explicitly in the probe metadata. In the benchmark survey, 88.1\% of
45,811 tasks had executable verdicts; judged tasks were concentrated in suites with
simulated users.

The builder receives the complete seed and must preserve its question and required outcome.
The identity check addresses semantic drift that executable controls alone can miss.
For example, a GAIA question about which scientist first predicted thinking machines
was transformed into an unrelated code-editing task. A matching checker and reference
could make that candidate executable without preserving the source task.

Admission is the conjunction
$\Lambda(s)=\lambda_{\mathrm{exec}}\wedge\lambda_{\mathrm{read}}\wedge
\lambda_{\mathrm{review}}$. The reviewer has read-only access and no shell; its verdict
requires a rationale grounded in the instruction and checker. Candidate construction
stops at either five rounds or 30 minutes. The time limit controls the cost of individual
rounds, which the round limit alone does not bound.

Let $\mathcal C=\mathcal C_1\times\cdots\times\mathcal C_d$ be the coverage space derived
from the campaign intent and seeds, and let $n(c)$ count admitted probes in cell $c$.
Exploration selects
\begin{equation}
c^\star\in\operatorname*{arg\,min}_{c\in\mathcal C}n(c),
\label{eq:explore}
\end{equation}
with uniform tie-breaking. Each new failure label queues three sibling builds, with
exploration forced after three consecutive verification builds. The coverage policy is
suited to campaigns with tens of expensive environment builds. For comparison, an
exploration term of the form $\sqrt{\ln N/n}$ is approximately 1.3 at $N=30,n=2$,
comparable to a reward bounded by one. This motivates the simpler allocation rule at
the scale studied here.

Sibling reproduction is based on the explained mechanism. Labels such as
\texttt{inconsistent-timing-artifact} and \texttt{invalid-timing-artifact} can describe the
same failure, so exact label equality is not used as a reproduction test. A sibling that
produces the mechanism in several rollouts still contributes one reproducing environment.

\subsection{Similarity calibration and induction}

Write $\mathcal D$ for the failures available to an induction run. For held-out evaluation,
select every fourth failure in corpus order to form the 25\% hold-out set
$\mathcal H$ before the pipeline and set $\mathcal D=\mathcal R\setminus\mathcal H$;
for corpus description, use the available corpus. The mean in
Equation~\eqref{eq:centre} is $\bar\phi=|\mathcal D|^{-1}\sum_{r\in\mathcal D}\phi(r)$.
Centering, calibration, grouping, and claim construction operate on $\mathcal D$.

Given calibrated thresholds $\tau_-<\tau_+$, pair decisions are
\begin{equation}
\mathrm{same}(r,r')=
\begin{cases}
1 & \mathrm{sim}(r,r')\geq\tau_+,\\
J(r,r') & \tau_-<\mathrm{sim}(r,r')<\tau_+,\\
0 & \mathrm{sim}(r,r')\leq\tau_- .
\end{cases}
\label{eq:band}
\end{equation}
In the ambiguous band, only pairs retained by the nearest-neighbor trim are judged;
unjudged band pairs introduce no graph edge. Calibration uses empirical judge agreement
on a stratified sample from the full similarity range:
\begin{align}
\tau_+&=\min\{\tau:\widehat{\Pr}(J=\mathrm{same}\mid\mathrm{sim}\geq\tau)\geq q_+\},
\nonumber\\
\tau_-&=\max\{\tau:\widehat{\Pr}(J=\mathrm{different}\mid\mathrm{sim}\leq\tau)\geq q_-\}.
\label{eq:calib}
\end{align}
Using the full range ensures that threshold estimation includes pairs outside the current
working band. The targets are $q_+=0.90$ and $q_-=0.98$.

For a component $K$ with at least two failures, the density statistics are
\begin{equation}
d_{\mathrm{in}}(K)=
\frac{1}{\binom{|K|}{2}}\sum_{\{r,r'\}\subseteq K}\mathrm{sim}(r,r'),
\qquad
d_{\mathrm{out}}(K)=
\max_{r\in K,\;r'\in\mathcal D\setminus K}\mathrm{sim}(r,r').
\label{eq:density}
\end{equation}
Edge weights are the centered cosine similarities, $w(r,r')=\mathrm{sim}(r,r')$,
on the retained edges. To split a violating component, sort its edges by increasing
weight and remove them one at a time until it disconnects. Recursively apply the
same test to each resulting component, using the remaining edges. This greedy
procedure approximates the grouping objective in Equation~\eqref{eq:cohesion}. When $K=\mathcal D$, no outside
failure exists; the implementation instead requires $d_{\mathrm{in}}(K)\geq\tau_+$.
Singletons are retained without a density test.

The boundary comprises the $k_{\mathrm{nn}}$ non-members maximizing
$\max_{r\in K}\mathrm{sim}(r,r')$. A generated claim is tested on every member and boundary
failure. The output includes the supporting failures, exclusions, and any failure to generate
a distinguishing claim.

\begin{algorithm}[!htbp]
\SetAlgoLined\DontPrintSemicolon
\KwIn{failures $\mathcal R$, judge $J$, margin $\mu$, precision targets $q_\pm$,
neighbor count $k_{\mathrm{nn}}$, hold-out fraction $h$}
\KwOut{groups, distinguishing claims, boundaries, held-out agreement}
Select every fourth failure for $\mathcal H$ ($h=0.25$); set $\mathcal D=\mathcal R\setminus\mathcal H$\;
Embed causes in $\mathcal D$, subtract their mean, and normalize\;
Calibrate $(\tau_-,\tau_+)$ on a stratified sample of pairs from $\mathcal D$\;
$E\leftarrow\{(r,r')\in\mathcal D^2:\mathrm{sim}(r,r')\geq\tau_+\}$\;
\ForEach{retained nearest-neighbor pair in the ambiguous band}{
  \lIf{$J(r,r')=\mathrm{same}$}{add $(r,r')$ to $E$}
}
$w(r,r')\leftarrow\mathrm{sim}(r,r')$ for each edge in $E$\;
$\mathcal K\leftarrow$ connected components of $(\mathcal D,E)$\;
\While{some non-singleton $K\in\mathcal K$ fails its density test}{
  remove edges of $K$ in ascending weight order until $K$ disconnects\;
  replace $K$ with the resulting connected components\;
}
\ForEach{$K\in\mathcal K$}{
  $B_K\leftarrow$ nearest $k_{\mathrm{nn}}$ non-members in $\mathcal D$\;
  $c_K\leftarrow$ claim covering $K$ and excluding $B_K$, or \textsc{unresolved}\;
}
Assign held-out failures using the claims; confirm assignments with $J$ against each selected group's first member\;
\Return{$\{(c_K,K,B_K)\}$ and held-out agreement}
\caption{Induction with held-out evaluation. For full-corpus description, omit the
hold-out split and membership evaluation.}
\label{alg:induce}
\end{algorithm}

\textbf{Held-out scoring.}
Let $a(r)$ be the group selected by the classification model after reading the
induced descriptions and the held-out failure, or $\bot$ if no description matches.
The classifier sees descriptions rather than the groups' member failures. For each
selected group $K$, let $r_K$ be its first member in corpus order. The reference
check is $J(r,r_K)$, using the same pairwise prompt as grouping, without supplying
the generated description.
For $\mathcal H_K=\{r\in\mathcal H:a(r)=K\}$, the implementation logs
\[
n_K=|\mathcal H_K|,\qquad
u_K=\sum_{r\in\mathcal H_K}J(r,r_K).
\]
These are the number of assignments and the number confirmed for each group;
held-out agreement counts confirmations among assignments. No-match outputs and
failed classification calls do not enter these counts. Both classification and
pairwise confirmation use the induction judge model. The reference labels
are therefore supplied by model-based pair comparisons.

\textbf{Embedding text.}
The serialization $\kappa(r)$ joins nonempty fields with spaces in this order:
summary, explanation, corrected action, trigger, and stage. Trigger and stage
are prefixed by ``Triggered by:'' and ``Stage:'' respectively. If all fields
are empty, the implementation falls back to the failure label.

\subsection{Hyperparameters}
\label{app:hyper}

Table~\ref{tab:hyper} lists the exposed settings. The corpus analysis uses $\mu=0.02$;
the component study uses $\mu=0.05$, and both settings are compared on the full corpus.
The thresholds $\tau_\pm$ are calibration outputs. The neighbor count controls the number
of ambiguous pairs judged and the number of boundary failures considered.
\begin{table}[!htbp]
\caption{Hyperparameters used in the experiments. The density-margin sweep is reported in Figure~\ref{fig:granularity}; fixed thresholds are used only in the corresponding ablation.}
\label{tab:hyper}
\small
\setlength{\tabcolsep}{4pt}
\zebrabody
\begin{tabularx}{\linewidth}{@{}clLr@{}}
\preprinttabletop
\headrow
\thead{Symbol} & \thead{Stage} & \thead{What it controls} & \thead{Value} \\
\preprinttablemid
$r_{\max}$ & \dt{admission} & rounds of build/verify/review before a candidate is discarded & 5 \\
$T_{\mathrm{build}}$ & \dt{admission} & wall-clock budget for one candidate, independent of rounds & 30\,min \\
$e$ & \dt{search} & verification builds before an exploration build is forced & 3 \\
$m$ & \dt{search} & siblings queued for a newly seen failure & 3 \\
$k$ & \dt{measurement} & rollouts per environment & 3 / 6 \\
$m_0$ & \dt{reproduction} & siblings required before a recurrence verdict is assigned & 3 \\
$\theta$ & \dt{reproduction} & fraction of siblings that must reproduce for \textsc{agent behaviour} & $2/3$ \\
$q_{+}$ & \dt{induction} & precision distance must reach before it may join a pair automatically & 0.90 \\
$q_{-}$ & \dt{induction} & precision distance must reach before it may separate a pair automatically & 0.98 \\
$\mu$ & \dt{induction} & density margin a component must clear to survive uncut & 0.02 \\
$k_{\mathrm{nn}}$ & \dt{induction} & neighbours kept per failure, for the judged band and for $B_m$ & 15 \\
$h$ & \dt{induction} & failures withheld from the whole pipeline & 0.25 \\
\totalrow\addlinespace[1pt]\arrayrulecolor{hair}\hline\addlinespace[1.5pt]
$\tau_{+},\tau_{-}$ & \dt{induction} & \dt{derived from $q_{\pm}$; fixed fallbacks $0.15$ and $-0.10$} & \soften{derived} \\
\preprinttablebottom
\end{tabularx}
\end{table}

\FloatBarrier
\section{Corpus and Failure Evidence}
\label{app:corpus}

\subsection{Agent configurations}

The panel contains 22 model--harness configurations, with one discovery campaign
per configuration. A 16-configuration campaign set
includes 13 models under Terminus~2, Claude Code with Sonnet~4.5, and Codex with
GPT-5.6-sol and GPT-5.4-nano. Six additional configurations support the matched design:
GPT-5.6-luna and GPT-5.6-terra under Terminus~2, and Sonnet~5 under Goose, opencode,
OpenHands, and Aider. Table~\ref{tab:profile} reports the 1,244 retained failures.

Campaign counts describe the observations collected for each configuration. For example,
Terminus~2 with Sonnet~4.5 contributes 143 failures in 52 groups, while Claude Code with
the same model contributes 37 failures in 16 groups. Such counts reflect the tasks
and depth reached during discovery; matched failure rates are reported separately in
Appendix~\ref{app:comparison}.

\ifprobeagentpreprint\else
\begin{table}[!htbp]
\caption{Corpus contributions by configuration. Groups includes all induced groups; shared counts those reached by more than one configuration. Largest is the total size of the largest group containing a failure from the configuration. Group counts are non-additive across rows.}
\label{tab:profile}
\small
\setlength{\tabcolsep}{4pt}
\zebrabody
\begin{tabularx}{\linewidth}{@{}Lrrrr@{}}
\preprinttabletop
\headrow
\thead{Configuration} & \thead{Records} & \thead{Groups} & \thead{Shared} & \thead{Largest} \\
\preprinttablemid
\ag{tb}{ds}{Terminus 2 / DeepSeek-V4-Pro} & 132 & \confirmed{68} & 7 & 8 \\
\ag{tb}{oa}{Terminus 2 / GPT-5.6-sol} & 132 & 58 & 4 & 8 \\
\ag{tb}{cl}{Terminus 2 / Claude Sonnet 5} & 105 & 56 & 5 & 4 \\
\ag{tb}{cl}{Terminus 2 / Claude Sonnet 4.5} & 143 & 52 & 7 & \confirmed{15} \\
\ag{tb}{ms}{Terminus 2 / Kimi K3} & 87 & 52 & 6 & 8 \\
\ag{tb}{xa}{Terminus 2 / Grok 4.5} & 84 & 47 & 3 & 5 \\
\ag{tb}{za}{Terminus 2 / GLM-5} & 72 & 35 & 0 & 4 \\
\ag{tb}{oa}{Terminus 2 / GPT-5.6-luna} & 64 & 33 & 4 & 8 \\
\ag{tb}{oa}{Terminus 2 / GPT-5.4-nano} & 73 & 30 & 3 & 12 \\
\ag{tb}{cl}{Terminus 2 / Claude Opus 5} & 55 & 24 & 1 & 4 \\
\ag{tb}{oa}{Terminus 2 / GPT-5.6-terra} & 50 & 21 & 2 & 6 \\
\ag{oh}{cl}{OpenHands / Claude Sonnet 5} & 40 & 20 & 2 & 5 \\
\ag{tb}{qw}{Terminus 2 / Qwen3.8-Max} & 39 & 20 & 5 & 5 \\
\ag{go}{cl}{Goose / Claude Sonnet 5} & 35 & 16 & 3 & 6 \\
\ag{cc}{cl}{Claude Code / Claude Sonnet 4.5} & 37 & 16 & 0 & 3 \\
\ag{tb}{mm}{Terminus 2 / MiniMax-M2.7} & 35 & 15 & 2 & \confirmed{15} \\
\ag{oc}{cl}{opencode / Claude Sonnet 5} & 30 & 14 & 2 & 6 \\
\ag{tb}{cl}{Terminus 2 / Claude Opus 4.8} & 11 & 8 & 2 & \confirmed{15} \\
\ag{cx}{oa}{Codex / GPT-5.6-sol} & 9 & 7 & 1 & 4 \\
\ag{tb}{za}{Terminus 2 / GLM-5.2} & 11 & 6 & 1 & 4 \\
\ag{cx}{oa}{Codex / GPT-5.4-nano} & \refuted{0} & \refuted{0} & 0 & 0 \\
\ag{ad}{cl}{Aider / Claude Sonnet 5} & \refuted{0} & \refuted{0} & 0 & 0 \\
\totalrow\addlinespace[1pt]\arrayrulecolor{hair}\hline\addlinespace[1.5pt]
\textbf{Corpus} & \textbf{1{,}244} & \textbf{565} & \textbf{27} & \textbf{15} \\
\preprinttablebottom
\end{tabularx}
\end{table}

\fi

\subsection{Seed coverage}
\label{app:seeds}

The seed tasks span retrieval, performance engineering, data pipelines, code editing,
terminal operation, tabular analysis, spreadsheets, clinical tasks, customer service, and
law. Table~\ref{tab:seeds} provides the recorded seed-level breakdown for the
16-configuration campaign set. Five entries account for approximately 94\% of the reported
group counts. Banking-support dialogue and data-pipeline repair contribute many observations,
consistent with their extended workflows and multiple opportunities for operational failure.

Seed eligibility also depends on executability and whether the target still fails the task.
Among twelve candidate tasks drawn from SWE-bench Verified and SWE-bench Pro, six were
solved by the target and six could not be built. Additional candidate suites failed to
produce usable seeds because their tasks were saturated, their runs produced no rewards,
or their reference solutions failed. These outcomes motivate task-level seed gates rather
than assuming every benchmark supplies suitable probes.

\begin{table}[!htbp]
\caption{Seed-level breakdown of the 16-configuration campaign set. Groups are counted within each seed and need not sum to a corpus-wide grouping. Configurations counts those contributing at least one failure under the seed.}
\label{tab:seeds}
\small
\setlength{\tabcolsep}{4pt}
\zebrabody
\begin{tabularx}{\linewidth}{@{}lLrrr@{}}
\preprinttabletop
\headrow
\thead{Suite} & \thead{Seed task} & \thead{Groups} & \thead{Records} & \thead{Configs.} \\
\preprinttablemid
\texttt{tau3-bench}     & \dt{banking\_knowledge task-081} & \confirmed{162} & 234 &  7 \\
\texttt{ade-bench}      & \dt{airbnb009}                   & 116 & 151 & 10 \\
\texttt{gaia}           & \dt{00d579ea}                    &  73 & 102 &  9 \\
\texttt{gaia}           & \dt{2a649bb1}                    &  72 &  96 &  9 \\
\texttt{ade-bench}      & \dt{asana005-hard}               &  57 &  80 & \confirmed{12} \\
\texttt{gso}            & \dt{pandas-9097263}              &   7 &  14 &  4 \\
\texttt{tau3-bench}     & \dt{retail-32}                   &   6 &  10 &  5 \\
\texttt{gso}            & \dt{numpy-893db31}               &   4 &  14 &  2 \\
\texttt{lawbench}       & \dt{2-4-23-zero-shot}            &   4 &   8 &  3 \\
\texttt{terminal-bench} & \dt{polyglot-c-py}               &   2 &   5 &  2 \\
\texttt{dabstep}        & \dt{1459}                        &   2 &   3 &  2 \\
\texttt{medagentbench}  & \dt{task10\_1}                   &   2 &   2 &  2 \\
\texttt{dabstep}        & \dt{17}                          &   1 &   1 &  1 \\
\preprinttablebottom
\end{tabularx}
\end{table}

\subsection{Attribution and breadth}
\label{app:evidence}

Table~\ref{tab:evidence} reports breadth and sibling evidence separately. A singleton
describes one observed failure; its attribution depends on the sibling tests linked to
the source failure. Cross-configuration membership describes breadth across agents; sibling tests
measure recurrence across task contexts.

Each test is keyed to the environment whose failure motivated its siblings. Across
146 tests, 311 failures and 161 groups inherit test evidence. Seven groups pool more than
one distinct test, and four contain tests with different verdict classes. Tests can
support several groups when trajectories from one tested environment yield failures that
induction separates. Within a group, each test contributes its reproduction count only
once. Of the 27 tests outside the 119-candidate experiment in Figure~\ref{fig:funnel},
14 have insufficient tested siblings. The remaining 404 groups have no linked sibling test.

\begin{table}[!htbp]
\caption{Evidence supporting the 565 induced groups. Breadth counts observed failures and configurations. Attribution pools distinct sibling tests associated with member failures; the two parts describe different dimensions of support.}
\label{tab:evidence}
\small
\zebrabody
\begin{tabularx}{\linewidth}{@{}Lrr@{}}
\preprinttabletop
\headrow
\thead{Breadth of supporting failures} & \thead{Groups} & \thead{Share} \\
\preprinttablemid
One failure & 361 & 63.9\% \\
Two or more failures, one configuration & 177 & 31.3\% \\
Two or more failures, spanning two configurations & 22 & 3.9\% \\
Two or more failures, spanning three or more & \confirmed{5} & 0.9\% \\
\totalrow\addlinespace[1pt]\arrayrulecolor{hair}\hline\addlinespace[1.5pt]
\textbf{Total} & \textbf{565} & \textbf{100\%} \\
\totalrow\addlinespace[3pt]
\headrow
\thead{Attribution from sibling tests} & \thead{Groups} & \thead{Share} \\
\preprinttablemid
Groups linked to sibling tests & \confirmed{161} & 28.5\% \\
\quad agent behavior & 33 & 5.8\% \\
\quad weak & 32 & 5.7\% \\
\quad task-specific & 53 & 9.4\% \\
\quad unverified & 43 & 7.6\% \\
\soften{Groups without sibling tests} & \soften{404} & \soften{71.5\%} \\
\preprinttablebottom
\end{tabularx}
\end{table}

\subsection{Example claims}

Table~\ref{tab:modes} presents four distinguishing claims.
The 565 induced groups comprise 361 single-failure groups, 177 multi-failure groups
from one configuration, and 27 groups spanning configurations. The largest group
contains 15 failures.
Overall, 29 of 565 groups are unresolved, including two of the 27 shared groups.
The other 536 groups have distinguishing claims.

\FloatBarrier
\section{Additional Induction Results}
\label{app:induction}

\subsection{Granularity}

\ifprobeagentpreprint
Section~\ref{sec:ablation} reports the repeated evaluations of the density constraint.
\else
\textbf{Repeated evaluations retain the benefit of the density constraint.}
On the 961-failure corpus, two evaluations of the full method yield held-out agreement
of 91.8\% and 91.4\%, compared with 78.0\% and 70.7\% without the constraint.
The full method retains higher agreement in both evaluations, supporting the
role of the density constraint in separating distinct errors. The additional
940-failure audit set yields 91.1\% agreement (Appendix~\ref{app:induction}).
\fi

With the same pair cache, grouping of the 961-failure corpus yields 505 groups:
381 singletons and 124 multi-failure groups. On the 940-failure audit set, induction
yields 491 groups\ifprobeagentpreprint.\else{} and 91.1\% held-out agreement.\fi

Figure~\ref{fig:granularity} reports the density-margin sweep on 961 failures. Without
the constraint, one component contains 650 failures. At the other extreme, $\mu=0.20$
produces 665 groups, of which only 38 contain multiple failures and none spans multiple
configurations. Intermediate settings retain more multi-failure groups while achieving
higher held-out agreement than unconstrained grouping.

On 1,244 failures, $\mu=0.05$ yields 659 groups, 75\% singletons, 13 shared groups,
a largest group of 10 failures, and 89.3\% held-out agreement.
At $\mu=0.02$, the corresponding values are 565, 64\%, 27, 15, and 88.5\%.
We select $\mu=0.02$ for its greater multi-failure and cross-configuration support.
Both settings are reported to show the resolution tradeoff.

\ifprobeagentpreprint\else
\begin{figure}[!htb]
\centering
\FigGranularity
\ifprobeagentpreprint
\par\vspace{5pt}
\FigCalibration
\fi
\caption{\ifprobeagentpreprint
\textbf{Grouping sensitivity and threshold calibration.}
(a) Groups with two or more failures and single-failure groups, with totals above the bars;
(b) held-out description agreement on 961 failures.
(c) Similarity ranges sent to the pairwise judge and the fraction labeled as matching.
Larger margins split groups; calibration concentrates judging on likely matches.
\else
\textbf{Effect of the density margin} on 961 failures. Top: all groups and groups
with at least two failures. Bottom: held-out membership agreement. Larger margins
progressively split groups and increase the singleton share.
\fi}
\label{fig:granularity}
\ifprobeagentpreprint\label{fig:calib}\fi
\end{figure}

\fi

\subsection{Comparison with a fixed taxonomy}

Figure~\ref{fig:taxonomy} compares induction with eight failure categories on 961 failures.
The taxonomy assigns 956 failures and leaves five unassigned. Its median category contains
94 failures, compared with a median group size of one under induction. The category
\texttt{incomplete} contains 264 failures and includes 163 induced groups.
The category-level group counts are non-additive. The median pairwise Jensen--Shannon
divergence between configurations is 0.21 in the taxonomy representation.

\subsection{Centering and calibration}

On 12 failures from three hand-labeled modes, centering increases the gap between
mean within-mode and between-mode cosine similarity from $0.048$ to $0.162$.
The respective means are $0.704$ and $0.656$ before centering, and $0.027$ and
$-0.135$ after centering. This gap measures how well similarity distinguishes the
labeled modes.

Figure~\ref{fig:calib}\ifprobeagentpreprint(c)\fi{} visualizes the fixed-band and calibrated-band comparison
reported in Section~\ref{sec:ablation}.
Across corpora of 664, 940, and 1,244 failures, the calibrated join thresholds are
0.630, 0.592, and 0.568, respectively.

\ifprobeagentpreprint\else
\input{figures/calibration}
\fi

\subsection{Pair selection and computational cost}

Section~\ref{sec:ablation} reports the neighbor-trimming and cache savings.
Of the 7,267 judged pairs on the 940-failure audit set, 2,711 are marked as matching.
The join threshold changes by 0.024 between these two runs. Cache reuse depends on
which pairs fall in the calibrated band, as well as which failure pairs are already stored.

\subsection{Human evaluation of failure descriptions}
\label{app:human-eval}

We randomly sample 30 held-out failures assigned to groups with generated
descriptions, keeping the descriptions fixed before annotation. Each failure is
paired with its assigned description and a similar alternative from another group,
selected by the maximum centered cosine similarity between the failure and any
member of that group. Two annotators independently evaluate all 60 pairs, yielding
120 judgments. Each item provides the task requirements, the specified failure,
relevant trajectory excerpts, checker feedback, and one candidate description;
the full trajectory is also available. Items are shuffled and presented separately,
with model and group identities, automatic judgments, and description type hidden.
Annotators decide whether the description accurately characterizes the failure,
answering yes, no, or insufficient information and giving a brief evidence-based
reason. A match requires the described error and key task conditions to agree
with the trajectory; different errors, incompatible conditions, or contradictory
evidence receive a no judgment. Unresolved evidence receives insufficient
information. Equivalent wording is accepted, while a shared broad outcome alone
does not establish a match. Both description types use the same criteria.
For each annotator, we divide yes judgments on assigned descriptions and no
judgments on alternatives by 30, then average the respective rates across
annotators. They accept 28/30 and 26/30 assigned descriptions and each reject
25/30 alternatives, giving mean acceptance and rejection rates of 90.0\% and
83.3\%, respectively.

\section{Agent Comparison and Resource Use}
\label{app:comparison}

\subsection{Controlled replay}

Replay fixes the environment pool and execution settings across configurations.
The pool contains exploration probes sampled evenly across source campaigns.
Sibling probes are excluded because they are constructed specifically to reproduce
a target's failure. Each probe retains its validated execution settings.
The replay pilot includes 20 environments, ten of which all tested agents fail.
Eight of the remaining environments form the discriminating pool.

Each configuration is scheduled for six rollouts in each of the eight environments.
Comparison uses five commonly measured environments and truncates each to the shallowest
available depth: $(6,6,4,1,1)$, totaling 18 matched trials. Environments that cannot
start under a harness, including network-isolated environments requiring a network
installation, are excluded from that common set. The Codex/GPT-5.4-nano configuration
has no valid replay executions and is excluded from failure-rate estimates.

Figure~\ref{fig:replay} reports point estimates and 95\% Wilson intervals.
Kimi~K3 under Terminus~2 has a failure rate of 22.2\%, and GPT-5.4-nano under
Terminus~2 has a failure rate of 94.4\%. The Opus~4.8/Opus~5 paired comparison has
discordant counts of six and one, with exact $p=0.13$.
The Sonnet and GLM version pairs yield $p=0.38$ and $p=0.45$.
For GPT-5.6-sol under Codex versus Terminus~2, $p=0.73$.

\ifprobeagentpreprint\else
\begin{figure}[!htb]
\centering
\FigReplay
\caption{\textbf{Matched replay results.} Rates use 18 matched trials per valid
configuration across five environments; intervals are 95\% Wilson intervals\ifprobeagentpreprint,
shown as tinted bands with each rate printed at its estimate\fi.
The pool was selected for discriminating failures. Infrastructure-only executions
are excluded.}
\label{fig:replay}
\end{figure}
\fi

\subsection{Configuration contrasts}

Figure~\ref{fig:pairs} displays selected model and harness contrasts using the group
counts in Table~\ref{tab:profile}. The model comparisons hold Terminus~2 fixed.
The harness comparisons hold Sonnet~4.5 or GPT-5.6-sol fixed.
These counts summarize the failures collected during discovery.

\ifprobeagentpreprint\else
\begin{figure}[!htb]
\centering
\FigPairs
\caption{\textbf{Group counts for selected configurations.}
(a) Model-version pairs under Terminus~2.
(b) Harness pairs with the same model. Counts match Table~\ref{tab:profile}.
\ifprobeagentpreprint
Grey and navy labels give the first and second configurations; the tinted band spans the
change.
\else
The pale and solid markers denote the first and second configurations, respectively.
\fi}
\label{fig:pairs}
\end{figure}
\fi

The model and harness comparisons hold four seed tasks,
builder, engine, intent, and rollout budget fixed.
GPT-5.6-luna, GPT-5.6-sol, and GPT-5.6-terra under Terminus~2 define
three model pairs. Sonnet~5 under Terminus~2, Goose, opencode, and OpenHands defines
six harness pairs.
Because environment generation remains adaptive, these campaigns study shared discoveries;
replay supplies the fixed-task performance comparison.

\ifprobeagentpreprint
Figure~\ref{fig:pairshare} in Section~\ref{sec:profile} summarizes the shared failure
modes. Within the nine related pairs, the three model pairs share 1, 2, and 0 groups;
the six harness pairs share 2, 1, and four zeros.
\else
Among the 190 pairs of failure-producing configurations, 30 share at least one group.
Four of the nine designed pairs share a group, compared with 26 of the other 181 pairs
(Figure~\ref{fig:pairshare}). The three model pairs share 1, 2, and 0 groups; the six
harness pairs share 2, 1, and four zeros.
\fi Mean overlap is 0.67 versus 0.19 groups per
pair at $\mu=0.02$, and 0.44 versus 0.07 at $\mu=0.05$.
Pairs reuse configurations, so these are descriptive comparisons rather than independent
replicates.

\ifprobeagentpreprint\else
\begin{figure}[!htb]
\centering
\FigPairShare
\caption{\textbf{Failure sharing across configuration pairs.}
\ifprobeagentpreprint
Bars show the fraction of pairs sharing at least one induced group: $4/9$ related
pairs and $26/181$ other pairs. Related pairs differ only in model version or harness.
\else
Each square represents a pair and is filled when the configurations share at least
one group. The designed pairs share at a rate of $4/9$; other pairs share at $26/181$.
\fi}
\label{fig:pairshare}
\end{figure}

\fi

The harness comparison also differs in observed failure count and timeout rate.
Terminus~2 with Sonnet~5 contributes 105 failures, versus 30--40 for each of the other
three harnesses, and times out on 46\% of rollouts versus 12--13\%.
It consumes 19.8M input tokens per rollout versus 3.2--8.6M.
Restricting shared support to completed rollouts changes the two nonzero harness-pair
overlaps from 2 and 1 to 1 and 0; the other overlaps remain zero.

\ifprobeagentpreprint\FloatBarrier\Needspace{10\baselineskip}\fi
\subsection{Token use}
\label{sec:cost}

Replay accounting is available for 15 configurations, each scheduled for 48 rollouts.
Mean input tokens per rollout range from 112k for Sonnet~5 under Terminus~2 to
3.49M for Sonnet~4.5 under Claude Code, a 31-fold spread. Among the 13 configurations
using Terminus~2, the spread is approximately tenfold.
Figure~\ref{fig:cost}\ifprobeagentpreprint(b)\fi{} relates token use over the replay pool to the matched failure rates.
For the Terminus~2 configurations, Spearman correlation is $\rho=0.50$ ($p=0.08$).
This association does not identify whether higher token use contributes to failure or
results from extended attempts on difficult tasks.

\ifprobeagentpreprint\else
\begin{figure}[!htb]
\centering
\FigCost
\caption{\textbf{Replay token use and matched failure rate.}
Input tokens per rollout use the replay accounting; failure rates use the 18 matched
trials. The horizontal axis is logarithmic.
\ifprobeagentpreprint
Numbers identify configurations in the key; marker color and shape identify
harnesses. The nearby MiniMax-M2.7 and DeepSeek-V4 points are labeled together.
\else
Model marks and harness rings identify configurations; nearby MiniMax-M2.7 and
DeepSeek-V4 values share a label.
\fi}
\label{fig:cost}
\end{figure}
\fi

Across the 16-configuration discovery set, target execution consumes 365M input and
7.3M output tokens over 936 rollouts with recorded usage. These counts include campaign
and seed-gate calls reaching the model; the execution audit separately examines all
trial outcomes.

\clearpage
\begingroup
\pretocmd{\subsection}{\Needspace{26\baselineskip}}{}{}
\section{Admission Analysis and Corpus Audit}
\label{app:validity}

\subsection{Admission-gate findings}
\label{app:gates}

The gates run sequentially and stop at rejection. Table~\ref{tab:gates} summarizes
1,451 build rounds. Execution rejects 349 candidates; semantic review rejects 638
candidates that passed the executable controls. The static source gate has no separately
recorded rejection. Inaccessible assertions are instead identified by the reviewer on
59 rounds. These are conditional rejection counts from the sequential pipeline.

\begin{table}[!htbp]
\caption{Admission rejections over 1,451 build rounds.}
\label{tab:gates}
\small\zebrabody
\begin{tabularx}{\linewidth}{@{}Lrl@{}}
\preprinttabletop
\headrow
\thead{Gate} & \thead{Rounds rejected} & \thead{Evaluation cost} \\
\preprinttablemid
Executed: reference and null controls & 349 & two rollouts \\
Read: static checker inspection & 0 & no rollout \\
Reviewed: separate read-only agent & 638 & one review pass \\
\preprinttablebottom
\end{tabularx}
\end{table}

\ifprobeagentpreprint
Section~\ref{sec:validity} and Table~\ref{tab:review-findings} report the semantic-review
findings. The gate counts in Table~\ref{tab:gates} show where these checks enter
probe admission.
\else
Section~\ref{sec:validity} discusses the most common findings and explains how shortcuts
can pass the executable controls. Table~\ref{tab:review-findings} gives the full breakdown,
including missing instructions, inaccessible assertions, and leaked solutions.
A round may receive several findings.
\fi

\ifprobeagentpreprint\else
\begin{table}[!htbp]
\caption{\ifprobeagentpreprint
Semantic-review findings over 1,451 build rounds. A round may receive multiple findings.
\else
Semantic-review findings. Categories may overlap within a build round.
\fi}
\label{tab:review-findings}
\small\zebrabody
\begin{tabularx}{\linewidth}{@{}Lr@{}}
\preprinttabletop
\headrow
\thead{Finding} & \thead{Rounds} \\
\preprinttablemid
A shortcut receives reward without completing the intended work & 565 \\
The checker does not evaluate the requested outcome & 323 \\
The checker imposes an unstated requirement & 255 \\
The instruction omits a rule needed to solve the task & 220 \\
The instruction cannot be followed as written & 92 \\
A task claim lacks an implementing mechanism & 81 \\
Checker assertions are inaccessible to the reviewer & 59 \\
The environment leaks its solution & 51 \\
\preprinttablebottom
\end{tabularx}
\end{table}

\fi

\subsection{Task and failure quality}

Table~\ref{tab:validity} summarizes the quality measurements.
The manual audit uniformly samples 20 environments and finds one hard error in an
expected value. Six of the 20 have either a wrong expected value or an ambiguity in
the required output format. Format defects include byte-exact checks on separator
spacing or key order that the instruction does not specify.
The per-failure classification uses the 940-failure audit set.

The answer-consistency audit finds that 2.2\% of 222 eligible variants change their
seed's expected answer. For judged environments, 225 reference replays across
72 environments have zero observed reward spread. All nine audited environments
failed by every tested agent accept their reference solutions. These controls assess
answer consistency, judge repeatability, and reference feasibility, respectively.

\begin{table}[!htbp]
\caption{Corpus quality audit. The manual sample contains 20 environments; failure classification covers 940 failures. Judge consistency measures repeatability of the reference verdict.}
\label{tab:validity}
\small
\setlength{\tabcolsep}{4pt}
\zebrabody
\begin{tabularx}{\linewidth}{@{}LLr@{}}
\preprinttabletop
\headrow
\thead{Check} & \multicolumn{1}{M}{\thead{Method}} & \thead{Result} \\
\preprinttablemid
Hard ground-truth errors        & \dt{unbiased sample of 20 environments, manual} & \refuted{5\%} \\
Wrong \emph{or} format-ambiguous & \dt{same sample} & 30\% \\
Variant overrules its seed's answer & \dt{automatic, all 222 eligible} & 2.2\% \\
\textbf{Records that are substantive} & \dt{per-failure classification, all 940} & \confirmed{\textbf{95.0\%}} \\
Records that are harness artifacts & \dt{same} & 3.1\% \\
Records that are format artifacts  & \dt{same} & 1.9\% \\
Judge self-consistency          & \dt{225 oracle replays over 72 judged environments} & spread $0.000$ \\
\textbf{All-fail environments still solvable} & \dt{reference solution re-run} & \confirmed{\textbf{9/9}} \\
\preprinttablebottom
\end{tabularx}
\end{table}

\endgroup
\end{document}